%% file: main.tex
\relax
\documentclass[10pt,twocolumn,letterpaper]{article}

\usepackage{cvpr}

\usepackage{graphicx}
\usepackage{amsmath}
\usepackage{amssymb}
\usepackage{booktabs}
\usepackage{graphicx}
\usepackage{amsmath}
\usepackage{amssymb}
\usepackage{multirow}
\usepackage{caption}
\usepackage{booktabs}
\usepackage{array}
\usepackage{pifont}
\usepackage{enumitem}
\usepackage{color}
\usepackage{amsmath}
\usepackage{arydshln} 
\DeclareCaptionFont{elevenpt}{\fontsize{9.8pt}{12pt}\selectfont #1}

\usepackage[pagebackref,breaklinks,colorlinks]{hyperref}

\usepackage[capitalize]{cleveref}

\newcommand{\Pc}{\ensuremath{\mathcal{P}}\xspace}

\newcommand{\Equation}[1]{Equation~\eqref{eq:#1}\xspace}
\newcommand{\Figure}[1]{Figure~\ref{fig:#1}\xspace}
\newcommand{\Table}[1]{Table~\ref{tab:#1}\xspace}
\newcommand{\PointPWC}{PointPWC~\cite{pointpwc}\xspace}
\newcommand{\FLOT}{FLOT~\cite{flot}\xspace}

\begin{document}

\title{SCTN: Sparse Convolution-Transformer Network for Scene Flow Estimation}

\author{\\Bing Li \quad Cheng Zheng \quad Silvio Giancola  \quad 
	Bernard Ghanem \\ 
	Visual Computing Center, KAUST, Thuwal, Saudi Arabia\\ \\}

	\maketitle
	
	\begin{abstract}
		We propose a novel scene flow estimation approach to capture and infer 3D motions from  point clouds. 
		Estimating 3D motions for point clouds is challenging,  since a point cloud is unordered  and its  density is significantly non-uniform. Such  unstructured data poses difficulties in  matching corresponding points between point clouds,  leading to inaccurate flow estimation.
		We propose a novel architecture  named Sparse Convolution-Transformer Network (SCTN) that  equips the 
		sparse convolution with the transformer. Specifically,
		by leveraging the sparse convolution, SCTN  transfers irregular point cloud into  locally  consistent flow features for estimating  continuous and consistent   motions within   an object/local object part.
		We further propose to explicitly learn point relations using a point transformer module,  different from exiting  methods.
		We show that the learned relation-based contextual information is rich and helpful for  matching corresponding points, benefiting scene flow estimation. 
		In addition,  a novel  loss function is proposed to adaptively encourage flow consistency according to feature similarity.  
		Extensive experiments demonstrate that our proposed approach achieves a new state of the art in scene flow estimation.  Our approach achieves an error of 0.038 and 0.037 (EPE3D) on FlyingThings3D and  KITTI Scene Flow respectively, which significantly outperforms previous
		methods by  large margins.
	\end{abstract}

	\section{Introduction}
	\label{sec:intro}

	Understanding 3D dynamic scenes is critical to many real-world applications such as autonomous driving and robotics.  Scene flow is  the 3D motion of points in a dynamic scene, which provides low-level  information for scene understanding \cite{vedula1999three,flownet3d,geiger2013vision}. 
	The estimation of the scene flow can be a building block for more complex applications and tasks such as 3D object detection~\cite{shi2020pv}, segmentation~\cite{thomas2019kpconv} and tracking~\cite{qi2020p2b}.
	However, many previous scene flow methods estimate the 3D motion  from stereo or RGB-D images.  With the increasing popularity of point cloud data, it is desirable to  estimate 3D motions  directly from 3D point clouds.

	\begin{figure}[t]
		\centering
		\begin{tabular}{m{0.2cm}<{\raggedright\arraybackslash}m{2.2cm}<{\centering}m{2.2cm}<{\centering}m{2.1cm}<{\centering}}
			{pc1}&\includegraphics[angle=-0, bb=0 0  619 559,trim={0.2cm 0 0 0.1cm},clip,width=0.123\textwidth]{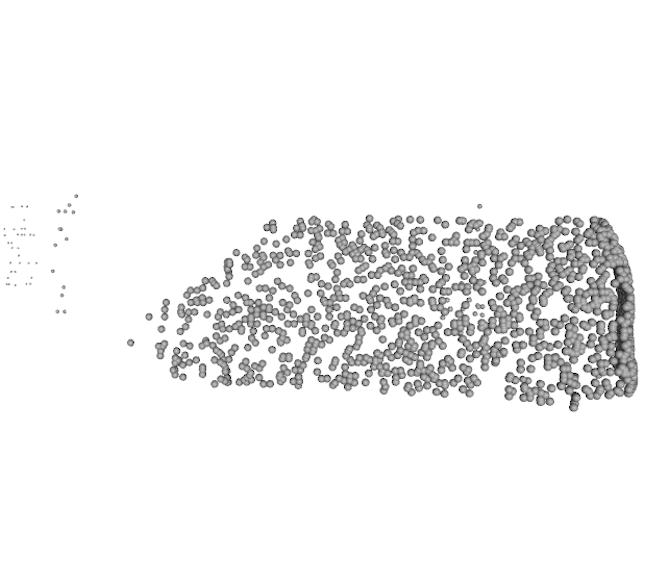}
			&
			\includegraphics[angle=-0, bb=0 0  619 559,trim={0.2cm 0 0 0.1cm},clip,width=0.123\textwidth]{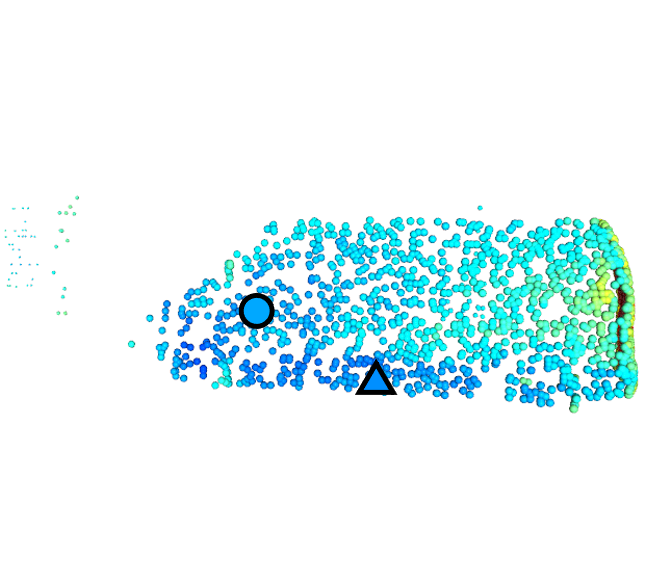} 
			& 	\includegraphics[angle=-0, bb=0 0  619 559,trim={0.2cm 0 0 0.1cm},clip,width=0.123\textwidth]{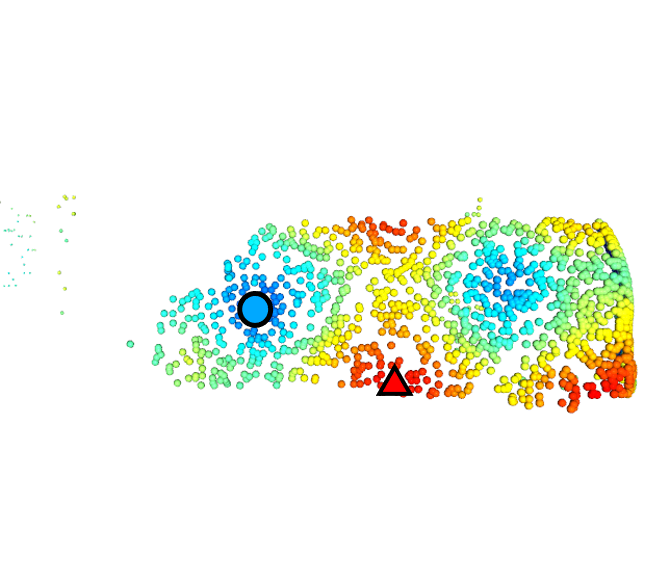}\\
			\hdashline[1.5pt/5pt]
			pc2&\includegraphics[angle=-0,bb=0 0  619 559,trim={0.2cm 0 0 0.1cm},clip,width=0.123\textwidth]{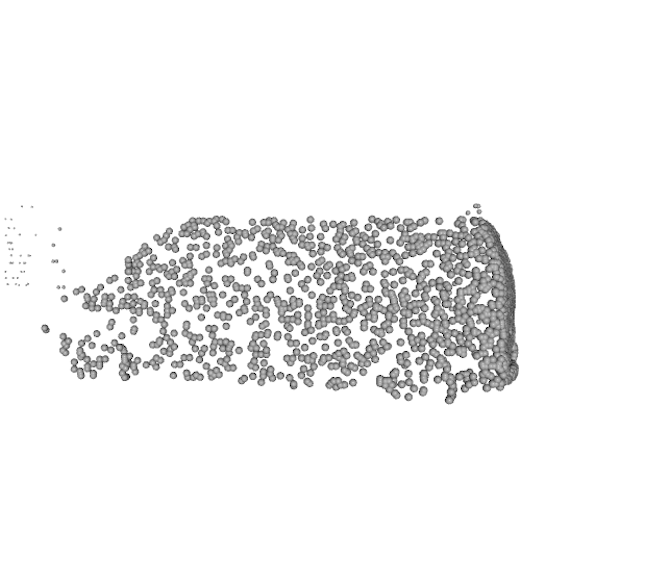}
			& 	\includegraphics[angle=-0, bb=0 0  619 559,trim={0.2cm 0 0 0.1cm},clip,width=0.123\textwidth]
			{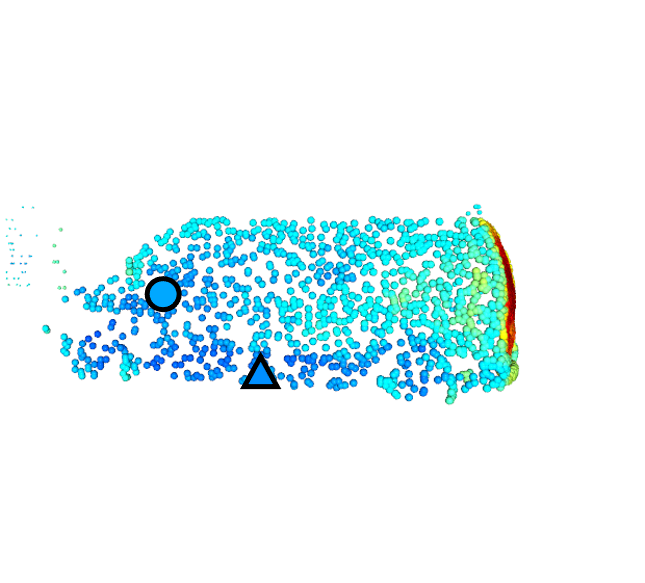} 
			& 	\includegraphics[angle=-0, bb=0 0  619 559,trim={0.2cm 0 0 0.1cm},clip,width=0.123\textwidth]
			{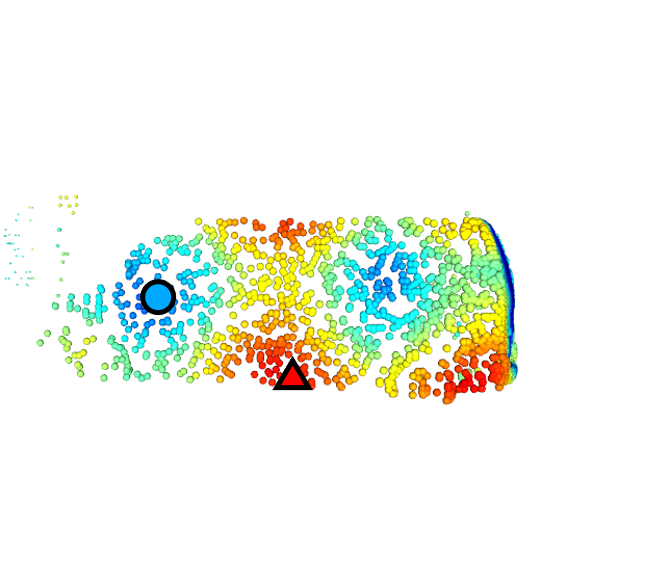}\\
			&(a) point clouds & (b) FLOT& (c) Ours\\
			
		\end{tabular}
		\caption{Illustrating the advantage of  our SCTN in feature extraction, where first and second rows  indicate the two point clouds,  (b) and (c) visualize  their  features  extracted by  FLOT and our SCTN, respectively. Circles $\bigcirc$ or  triangles $\bigtriangleup$ in (b)(c) indicate a pair of corresponding points between the  point clouds, respectively. $\bigcirc$ and  $\bigtriangleup$ are not corresponding to each other, however, their features extracted by FLOT are improperly similar, which are less discriminative and would lead to inaccurate predicted flows.  In contrast, our SCTN extracts locally consistent while discriminative features.   }
		\label{fig:fig1}
	\end{figure}

	Recent methods \eg \cite{pointpwc,flownet3d,wei2020pv,flot,FESTA,HCRFFlow}    propose deep neural networks to learn scene flow from point clouds in an end-to-end way,  which  achieves  promising estimation performance.  However,  estimating scene flow from point clouds   is still challenging.
	In particular,
	existing methods 
	\cite{flownet3d,flot} extract feature for  each point by  aggregating information  from its local neighborhood. However,  such extracted features are not discriminative enough to matching corresponding points between point clouds (see Figure \ref{fig:fig1}), leading to inaccurate flow estimation, since the feature extraction of these methods ignore two facts.
	First,    the  density of points is significantly non-uniform within a  point cloud. It is non-trivial  to  learn  point features  that are   simultaneously favorable for both  points from   dense regions  and  those from sparse regions.
	Second, due to Lidar and object motions, the density of points within an object often varies  at the temporal dimension,  leading that the geometry patterns of corresponding local regions are inconsistent between consecutive point clouds. As a result, extracted features, that are aggregated from only  local regions without modeling point relations, is insufficient for scene flow estimation.

	Another challenge lies in that  most previous methods have to  train a model on the synthetic
	dataset to estimate scene flow for real-world data. However, there exists domain shift between
	the synthetic dataset and the real-world one.
	For example, most objects in the synthetic dataset are rigid  and undergo rigid motions, while real-world data contains many non-rigid objects whose motions   are  consistent  within local object parts, rather than the whole object.  Consequently, the performance of the trained model is degraded when handling real-world data.  Yet, most  recent work \cite{flownet3d,flot,pointpwc} do not explicitly constrain the estimated flows.
	We would like to enforce  the predicted flows in a local region of an object to be \textit{consistent} and \textit{smooth} in 3D space.

	In this work, we resort to feature representations and loss functions to estimate accurate scene flow  from point clouds. 
	In particular, we explore novel feature representations (information) that would help to infer an accurate and locally consistent scene flows. We therefore propose a Sparse Convolution-Transformer Network (SCTN) which incorporates the merits of dual feature representations provided by our two proposed modules. In particular, to address the issue of spatially non-uniform and temporally varying density of dynamic point clouds, we  propose a Voxelization-Interpolation based Feature Extraction (VIFE) module. VIFE  extracts features from  voxelized point clouds rather than original ones,  and then interpolates features for each point, which encourages to generate locally  consistent flow features. To furture improve discriminability of extracted features from the VIFE module, we propose to additionally model  point relations in the feature space, such that extracted features  capture important contextual information.
	Inspired by impressive performance of transformer in object detection tasks \cite{DETR2020}, we propose a Point Transformer-based Feature Extraction (PTFE) module to explicitly learn point relations based on  transformer  for  capturing complementary information.

	In addition, we propose a spatial consistency loss function with  a new architecture that equips stop-gradient for training. The loss adaptively controls the flow consistency according to the similarity of point features.  
	Our experiments demonstrate that our method significantly outperforms state-of-the art approaches on standard scene flow datasets: FlyingThings3D \cite{flythings3d} and KITTI Scene Flow \cite{Menze2018JPRS}.

	\vspace{3pt}
	\noindent\textbf{Contributions.} Our contributions are fourfold:

	\begin{enumerate}[noitemsep,topsep=0pt ] 
		\item Instead of designing convolution kernels, our VIFE module leverages simple  operators -- voxelization and interpolation for feature  extraction, showing such smoothing operator is effective to extract local consistent while discriminative  features  for scene flow estimation.   
		
		\item  Our PTFE module shows that  explicitly modeling
		point relations  can provide rich contextual information
		and is helpful for matching corresponding points,
		benefiting scene flow estimation. We are the first to introduce transformer for scene flow estimation. 
		
		\item We propose a new consistency loss equipping stop-gradient-based architecture that helps
		the model trained on synthetic dataset well adapt to real
		data, by controlling spatial consistency of estimated flows.

		\item We propose a novel network that outperforms the state-of-the-art methods with remarkable margins on both FlyingThings3D and KITTI Scene Flow  benchmarks.
		
	\end{enumerate}

	\begin{figure*}[t]
		\centering
		\includegraphics[bb=0 0  3984 980,width=1\textwidth]{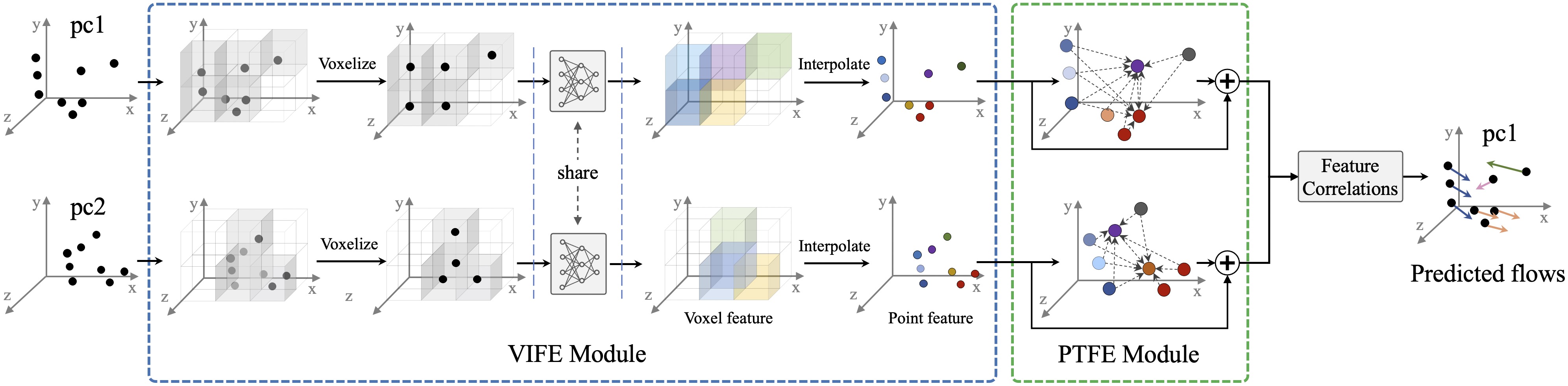}
		\caption{{Overall framework of our SCTN approach.}  
			Given two consecutive point clouds,  the Voxelization-Interpolation Feature Extraction (VIFE)  extracts features from voxelized point clouds and then projects back the voxel features into point features.
			These point features are fed into the Point Transformer Feature Extraction (PTFE) module to explicitly learn point relations.
			With fused  features of VIFE and PTFE module,   SCTN  computes point correlations  between the point clouds and predicts flows.
		}
		\label{fig:framework}
	\end{figure*}

	\section{Related Work}

	\textbf{Optical Flow}. Optical flow estimation is defined as the task of predicting the pixels motions between consecutive 2D video frames.
	Optical flow is a fundamental tool for 2D scene understanding, that have been extensively studied in the literature.
	Traditional methods \cite{horn1981determining,black1993framework,zach2007duality,weinzaepfel2013deepflow,brox2009large,ranftl2014non} address the problem of estimating optical flow as an energy minimization problem, that does not require any training data.
	Dosovitskiy~\etal~\cite{dosovitskiy2015flownet} proposed a first attempt for an end-to-end model to solve optical flow based on convolution neural network (CNN). Inspired by this work, many CNN-based studies have explored data-driven approaches for optical flow \cite{dosovitskiy2015flownet,mayer2016large,ilg2017flownet,hui2018liteflownet,hui2020liteflownet3,PWCNet,teed2020raft}.

	\textbf{Scene Flow from Stereo and RGB-D Videos.}
	Estimating scene flow from stereo videos have been studied for years~\cite{chen2020consistency,vogel2013piecewise,wedel2008efficient,ilg2018occlusions,jiang2019sense,teed2020raft-3d}.
	Many works estimate scene flow by jointly estimating stereo matching and optical flow from consecutive stereo frames \cite{MIFDB16}. Similar to optical flow, traditional methods formulate scene flow estimation as an energy minimization problem~\cite{huguet2007variational,wedel2008efficient}. Recent works estimate scene flow from stereo video using neural networks \cite{chen2020consistency}. For example, networks for disparity estimation and optical flow are combined in \cite{ilg2018occlusions,ma2019drisf}.
	Similarly, other works \cite{quiroga2014dense,sun2015layered} explore scene flow estimation from RGB-D video.

	\textbf{Scene Flow on Point Clouds.} Inspired by FlowNet~\cite{dosovitskiy2015flownet}, FlowNet3D~\cite{flownet3d} propose an end-to-end network to estimate 3D scene flow from raw point clouds. Different from traditional methods \cite{dewan2016rigid,ushani2017learning}, FlowNet3D~\cite{flownet3d} is based on PointNet++~\cite{pointnetplus}, and propose a flow embedding layer to aggregate the information from consecutive point clouds and extract scene flow with convolutional layers.
	FlowNet3D++~\cite{FlowNet3Dplus} improves the accuracy of FlowNet3D by incorporating geometric constraints. HPLFlowNet~\cite{HPLFlowNet} projects point clouds into permutohedral lattices, and then estimates scene flow using Bilateral Convolutional Layers. 
	Inspired by the successful optical flow method PWC-Net~\cite{PWCNet}, PointPWC \cite{pointpwc} estimates scene flow in a coarse-to-fine fashion, introducing cost volume, upsampling, and warping modules for the point cloud processing. 
	The most related recent work to our approach is FLOT \cite{flot}.
	FLOT addresses the scene flow as a matching problem between corresponding points in the consecutive clouds and solve it using optimal transport.
	Our method differs from FLOT~\cite{flot} in two aspects. 
	First, our method explicitly explores more suitable feature representation that facilitate the scene flow estimation.  
	Second, our method is trained to enforce the consistency of predicted flows for local-region points from the same object, which is ignored in FLOT \cite{flot}.
	Recently, Gojcic~\etal~\cite{weaklyrigidflow} explore weakly supervised learning for scene flow estimation using labels of ego motions as well as ground-truth foreground and background masks. Other works~\cite{pointpwc,flowstep3d,Mittal_2020_CVPR} study unsupervised/self-supervised learning for scene flow estimation on point clouds,
	proposing regularization losses that enforces local spatial smoothness of predicted flows.
	These losses are directly constraining points in a local region to have similar flows, but are not feature-aware.

	\textbf{3D Deep Learning.} 
	Many works have introduced deep representation for point cloud classification and segmentation~\cite{pointnetplus,thomas2019kpconv,zhang2019shellnet,liu2020closer,wu2019pointconv,li2018pointcnn,lei2020spherical,wei_hu,hengshuangTransformer}.
	Qi~\etal~\cite{pointnet} propose PointNet that learns point feature only from point positions. PointNet++ \cite{pointnetplus} extends PointNet by aggregating information from local regions. Motivated by PointNet++, many works~\cite{zhang2019shellnet,thomas2019kpconv,lei2020spherical,li2018pointcnn} design various local aggregation functions for point cloud classification and segmentation.
	Different from these point-based convolutions, \cite{wu20153d,WangLGST17} transform point cloud into voxels, such that typical 3D convolution can be applied. 
	However, such voxel-based convolution suffers from expensive computational and memory cost as well as information loss during voxelization. Liu~\etal~\cite{pvcnn} combine PointNet \cite{pointnet} with voxel-based convolution to reduce the memory consumption. For the sake of efficient learning, researchers have explored sparse convolution for point cloud segmentation \cite{PointContrast2020,minkowskiSPC} which shows impressive performance. Tang~\etal~\cite{spvnas} propose to combine PointNet with sparse convolution for large-scale point cloud segmentation.
	Differently,  we not only leverage  voxelization and interpolation  for feature extraction, but also explicitly model point relations to provide complementary information.

	\section{Methodology}
	\textbf{Problem Definition.}
	Given two consecutive point clouds  $\mathcal{P}^{t}$ and $\mathcal{P}^{t+1}$, scene flow estimation is to predict the 3D motion flow  of  each point from $\mathcal{P}^{t}$ to $\mathcal{P}^{t+1}$.   Let  $p^t_i$ be the 3D coordinates of $i$-th  point in $\Pc^{t} = \{p^t_i\}_{i=1}^{n_\Pc}$.  Like previous work \cite{flot}, we predict scene flow based on the correlations of  each point pair between  $\mathcal{P}^{t}$ and $\mathcal{P}^{t+1}$. 
	Given  a pair of points  $p_i^t\in \mathcal{P}^{t} $ and $p_j^{t+1} \in \mathcal{P}^{t+1}$, the correlation of the two points  is computed as follows:
	
	\begin{equation}
		{C}(p_i^t,p_j^{t+1}) =  \frac{(\mathbf{F}^{t}_i)^T \cdot \mathbf{F}^{t+1}_j}{\|\mathbf{F}^{t}_i\|_2\|\mathbf{F}^{t+1}_j\|_2}
		\label{eq:correlation}
	\end{equation}
	where   $\mathbf{F}^{t+1}_j$ is  the feature of    $p_j^{t+1}$, and $\|\cdot\|_2$ is the L2 norm.
	
	Point feature $\mathbf{F}$ is the key to computing the correlation  ${C}(p_i^t,p_j^{t+1})$  which further plays an important role in scene flow estimation.
	Hence, it is desirable to extract effective point features that enable point pairs in corresponding regions to achieve higher correlation values between $\mathcal{P}^{t}$ and $\mathcal{P}^{t+1}$.
	However, different from point cloud segmentation and classification that focus on static point cloud, scene flow estimation operates on dynamic ones, which  poses new challenges for feature extraction in two aspects. For example, the input point clouds of scene flow are not only irregular and unordered, but also its density is  spatially non-uniform and temporally varying, as discussed in previous sections.

	Our goal is to extract locally consistent while discriminative features for points, so as to  achieve accurate flow estimation. Different from exiting methods  directly extracting point feature from the neighborhood in  original point cloud, we 
	proposed two feature extraction modules for scene flow estimation. The    Voxelization-Interpolation based Feature Extraction (VIFE) is proposed to address the issue of point cloud's non-uniform density.
	With the features extracted by VIFE, Point Transformer based Feature Extraction (PTFE) further enhance feature discriminability by modeling point relations globally.
	Since the two kinds of  features  provide complementary information, we  fuse these features,  such that the fused features provide
	proper correspondences to predict accurate scene flows.

	\textbf{Overview.} Figure~\ref{fig:framework} illustrates the overall framework of our approach, that takes two consecutive point clouds $\mathcal{P}^{t}$ and $\mathcal{P}^{t+1}$ as inputs and predict the scene flow from $\mathcal{P}^{t}$ to $\mathcal{P}^{t+1}$.
	First, the VIFE module extracts  features for voxel points from voxelized point clouds, and  projects back the voxel features  into the original 3D points to  obtain locally consistent  point features.
	Second, our PTFE module improves the point feature representation by modeling relation between points.   
	Third, with features extracted by VIEF and PTFE module, we calculate the correlations of points between $\mathcal{P}^{t}$ and $\mathcal{P}^{t+1}$, where a sinkhorn algorithm \cite{flot,cuturi2013sinkhorn,chizat2018scaling} is leveraged to predict the flows.
	We train our method with an extra regularizing loss to enforce spatial consistency of predicted flows.

	\subsection{Voxelization-Interpolation based feature extraction} 
	\label{subsec:spconv}

	As mentioned in previous sections, the density of consecutive point clouds is spatially non-uniform and temporally varying, posing difficulties in feature extraction.  
	To address the issue, we  propose the Voxelization-Interpolation based Feature Extraction (VIFE) module. VIFE first voxelizes the consecutive point clouds into voxels. As illustrated in Figure \ref{fig:framework},  the spatial non-uniform distributions and temporally variations of points are reduced to some extent. 
	
	After that, VIFE conducts convolutions on voxel points rather than  all points of the point cloud, and then interpolate features for each point.    We argue that such simple operators \ie voxelization and interpolation,  ensure points in a local neighborhood to have smoother features, ideally leading to consistent flows in space.

	\textbf{Voxel feature extraction.} 
	We then leverage a U-Net \cite{ronneberger2015u} architecture network to extract feature from voxelized point clouds,  where convolution can be many types of  point cloud convolutions  such as pointnet++ used in FLOT \cite{flot}.  Here, we adopt sparse convolution \eg Minkowski Engine~\cite{minkowskiSPC}  for efficiency. More details are available in our supplementary material.

	\textbf{Point feature interpolation.} We project back the voxel features into point feature $\mathbf{F}^S_i$ for point $p_i$. In particular, we interpolate the point features from the $K$ closest voxels following equation \eqref{eq:interpolation}.
	$\mathcal{N}^v(p_i)$ represents the set of $K$ nearest neighboring non-empty voxels for the point $p_i$, $\mathbf{v}_k \in \mathbb{R}^C$ represents the feature of $k$-th closest non-empty voxel and $d_{ik}$ the Euclidian distance between the point $p_i$ and the center of the $k$-th closest non-empty voxel.

	\begin{equation}
		\mathbf{F}^S_i =  \frac{\sum_{k\in  \mathcal{N}^v(p_i)} d_{ik}^{-1}\cdot\mathbf{v}_k }{\sum_{k\in \mathcal{N}^v(p_i) }d_{ik}^{-1}}
		\label{eq:interpolation}
	\end{equation}
	
	We observed that close points are encouraged to have similar features, which  helps our method  generate consistent flows  for these points.  This is  favorable for local object parts or rigid objects with dense densities and consistent flows (\eg LiDAR points on a car at close range).

	\subsection{Point transformer based feature extraction} \label{subsec:transformer}
	Our VIFE module adopt  aggressive downsampling to obtain a large receptive field and low computation cost.  However,   aggressive downsampling  inevitably loses some important information \cite{spvnas}. In such case, the features of points with large information loss are  disadvantageous for estimating  their scene flow.
	To address this issue, we explicitly exploit point relations as a complementary information on top of the point feature extracted with VIFE. 
	Recent work \cite{zhu2020deformable,Zhao_2020_CVPR,DETR2020} employ transformer and self-attention to model internal relation in the features, achieving impressive performance in image tasks such as detection and recognition. 
	Similar trend appeared in point cloud classification and segmentation~\cite{ptc2021,hengshuangTransformer,nicoTransformer} showing the effectiveness of transformer in 3D.
	Inspired by these work, we resort to transformer for capturing point relation information as the point feature.

	In an autonomous navigation scenario,  point clouds represent complete scenes, with small object such as cars and trucks, but also large structure such as buildings and walls.
	The scene flows in such a large scene do not only depend on the aggregation from a small region, but rather a large one.
	As a results, we refrain in building a transformer for the local neighborhood as it would restrict the receptive field  or require deeper model (\ie increase the memory).
	Instead, our transformer module learns the  relation of each point to all other points, such that the transformer can adaptively capture the rich contextual information from a complete scene.

	Formally, given a point $p_i$, we consider every points in \Pc as query and key elements.
	Our transformer module builds a point feature representation for $p_i$ by adaptively aggregating the features of all points based on self-attention:
	\begin{equation}
		\mathbf{F}^R_i=\sum_{j=1}^{n_\Pc}{A}_{i,j} \cdot g_v(\mathbf{F}^{S}_j, \mathbf{G}_j)
	\end{equation}
	where $g_v$ is the a learnable function (\eg linear function), ${A}_{i,j}$ is an attention defining a weight of $p_j$ to $p_i$, $\mathbf{G}_j$ is the positional encoding feature of $p_j$.

	As pointed in literature \cite{Zhao_2020_CVPR,DETR2020}, the positional encoding feature can provide important information for the transformer. 
	The position encoding in recent transformer  work \cite{hengshuangTransformer} encodes the  \textit{relative} point position to neighbors for point cloud classification or segmentation. Different from those tasks, the task of scene flow is to find correspondences between consecutive point clouds. Thus, we argue that an \textit{absolute} position provides sufficient information to estimate the scene flow.
	Therefore, given a point $p$,  our  position encoding function encodes its \textit{absolute} position $p_j$:
	\begin{equation}
		\mathbf{G}_j = \phi(p_j)
	\end{equation}
	where  $\phi$ is a MLP layer. Using absolute positions reduce computational cost, compared with using relative positions.
	
\begin{figure}[t]
	\centering
	\includegraphics[width=0.65\linewidth, keepaspectratio]{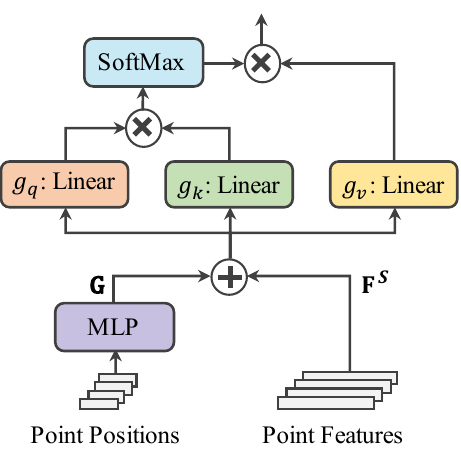}
	\caption{{Details of our point transformer based feature extraction module (PTFE).}
		$\otimes$ and $\oplus$ correspond to matrix multiplication and addition operations, respectively.}
	\label{fig:transformer}
\end{figure}
	
	We calculate an attention  ${A}_{i,j}$ as the similarity between the features of $p_i$ and $p_j$ in an embedding space.
	The similarity is estimated using features and position information:
	
	\begin{equation}
		{A}_{i,j}  \propto \exp(\frac{(g_q(\mathbf{F}^{S}_i,\mathbf{G}_i))^T \cdot g_k(\mathbf{F}^{S}_j,\mathbf{G}_j)}{c_a})
	\end{equation} 
	where $g_q(\cdot,\cdot)$ and $g_k(\cdot,\cdot)$ are the learnable mapping functions to project feature into an embedding space, and $c_a$ is the output dimension of $g_q(\cdot,\cdot)$ or $g_k(\cdot,\cdot)$. $A_{i,j}$ is further normalized such that $\sum_j{A}_{i,j}=1$.
	The architecture of our transformer module is illustrated in \Figure{transformer}.

	\subsection{Flow prediction}\label{subsec:sinkhorn}

	Since the point feature from our VIFE  and  PTFE modules provide complementary information, we fuse the two kinds of features through skip connection for each point, \ie $\mathbf{F}_i= \mathbf{F}^S_i + \mathbf{F}^R_i$. By feeding the fused point features   into Eq. \ref{eq:correlation}, we compute the  correlations   of all pairs $\mathbf{C}(\mathcal{P}^{t}, \mathcal{P}^{t+1})=\{C(p_i^t,p_j^{t+1})\}$ between the two consecutive point clouds. 
	
	With the estimated point correlations, we  adopt the Sinkhorn algorithm to estimate soft correspondences   and  predict flows  for $\mathcal{P}^{t}$, following FLOT.

	\subsection{Training losses}\label{subsec:loss}
	
	We train our model to regress the scene flow in a supervised fashion, on top of which we propose a Feature-aware Spatial Consistency loss, named ``FSC loss'', that enforces similar features to have similar flow.
	The FSC loss provides a better generalization and transfer capability between the training and testing datasets.

	\textbf{Supervised loss.} We define in \Equation{suploss} our supervised loss $E^s$ that minimize the $\mathit{L}_1$-norm difference between the estimated flow and the ground truth flow for the non-occluded points. 
	$\mathbf{u}^*_i$ and $\mathbf{u}_i $ are respectively the ground-truth and predicted motion flow for the point $p_i \in \Pc$ and $m_i$ is a binary indicator for the non-occlusion of this point, \ie $m_i=0$ if $p_i$ is occluded, otherwise $m_i=1$.

	\begin{equation}
		E^s = \sum_{i}^{N} m_i \| \mathbf{u}_i - \mathbf{u}^*_i \|
		\label{eq:suploss}
	\end{equation}

	\begin{figure}[t]
		\centering
		\includegraphics[width=1.0\linewidth, keepaspectratio]{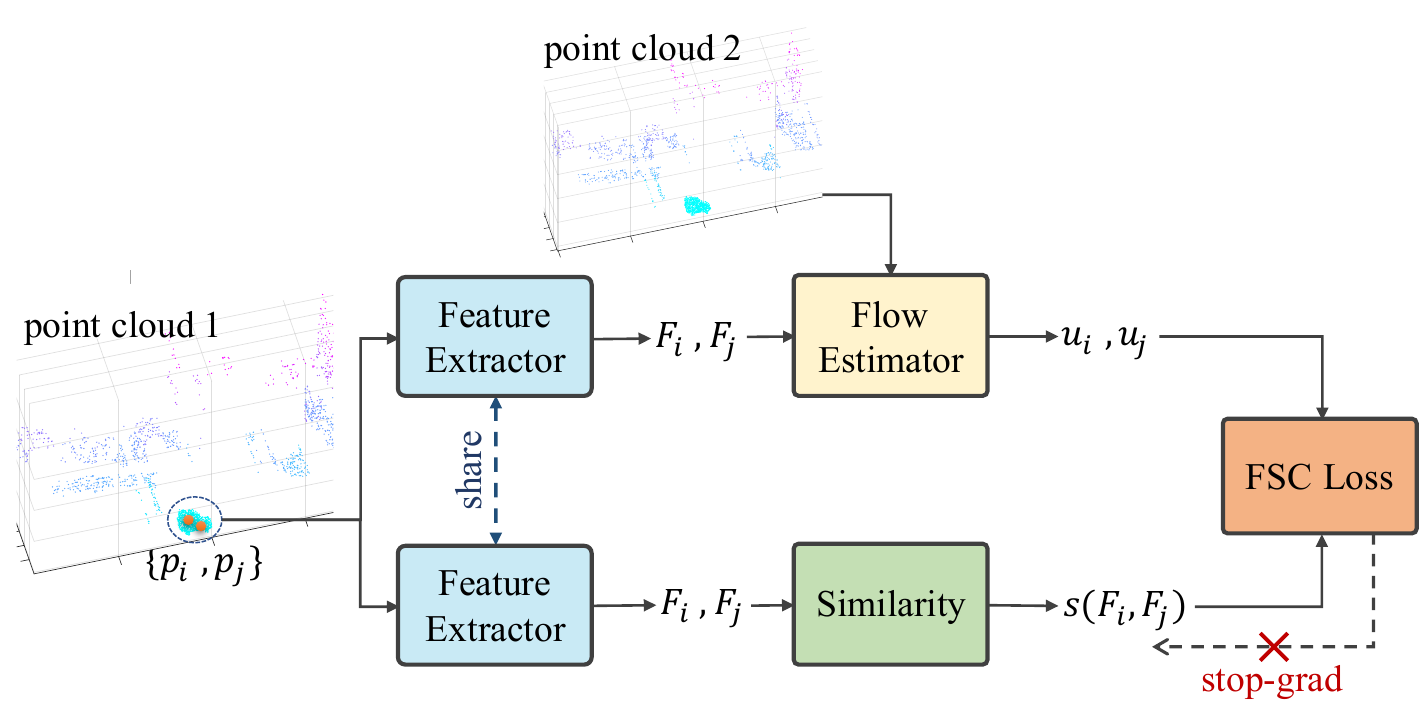}
		\caption{{Stop gradient for the FSC loss.}
			We extract the flow from a neighboring point as well as the similarity between their features. We optimize the FSC loss without back-propagating the gradients to the features from the similarity branch to avoid degenerate cases.
		}
		\label{fig:consistencyLoss}
	\end{figure}

	\textbf{Feature-aware Spatial Consistency (FSC) loss.} Given a local region, points from the same object usually has  consistent motions, resulting in similar flows. To model such phenomena, we propose a consistency loss that ensures points within an object/local object part to have similar predicted flows. Yet, object annotations are not necessarily available.  Instead,  we propose to control flow consistency according to feature similarity.
	That is, given a local region, if two points are of larger feature similarity, they are of the higher probability that belongs
	to the same object.
	In particular, given a point $p_i$ with predicted flow $\mathbf{u}_i$ and its  local neighborhood $\mathcal{N}(p_i)$, we enforce the flow $\mathbf{u}_j$ of  $p_j \in \mathcal{N}(p_i)$  to be  similar to $\mathbf{u}_i$,  if the feature $\mathbf{F}_i$ of $p_j$  is similar to $\mathbf{F}_j$ of $p_j$.
	Formally, we define the FSC loss  as follows:
	\begin{equation}
		E^c =\sum_{i=1}^N \frac{1}{K}\sum_{p_j\in \mathcal{N}(p_i)} s(\mathbf{F}_i,\mathbf{F}_j) \cdot \| \mathbf{u}_i-\mathbf{u}_j\|_2  
		\label{eq:simense}
	\end{equation}
	where the similarity function $s(\mathbf{F}_i,\mathbf{F}_j)$ of $p_i$ and $p_j$ is defined as  $1-\exp(-(\mathbf{F}_i)^T\cdot\mathbf{F}_j/\tau)$ with $\tau$ being a temperature hyper-parameter, $K$ is the number of points in $\mathcal{N}(p_i)$.

\begin{table*}[t]
	\centering
	\small
	\begin{tabular}{p{2.3cm} p{5.5 cm}  p{1.9cm}<{\centering} p{1.9cm}<{\centering} p{1.8cm}<{\centering}p{1.9cm}<{\centering} }
		\toprule
		Dataset & Method & EPE3D(m) $\downarrow$ & Acc3DS $\uparrow$ & Acc3DR $\uparrow$ & Outliers $\downarrow$ \\
		\midrule
		\multirow{5}{*}{FlyingThings3D} 
		& FlowNet3D \cite{flownet3d}        &   0.114 &   0.412 &   0.771 &   0.602 \\
		& HPLFlowNet \cite{HPLFlowNet}      &   0.080 &   0.614 &   0.855 &   0.429 \\
		& PointPWC \cite{pointpwc}          &   0.059 &   0.738 &   0.928 &   0.342 \\
		& EgoFlow \cite{tishchenko2020self} &   0.069 &   0.670 &   0.879 &   0.404 \\
		& FLOT  \cite{flot}                 &   0.052 &   0.732 &   0.927 &   0.357 \\
		& SCTN (ours)                              &\bf0.038 &\bf0.847 &\bf0.968 &\bf0.268 \\
		\midrule
		\multirow{5}{*}{KITTI} 
		& FlowNet3D \cite{flownet3d}        &   0.177 &   0.374 &   0.668 &   0.527 \\
		& HPLFlowNet \cite{HPLFlowNet}      &   0.117 &   0.478 &   0.778 &   0.410 \\
		& PointPWC \cite{pointpwc}          &   0.069 &   0.728 &   0.888 &   0.265 \\
		& EgoFlow \cite{tishchenko2020self} &   0.103 &   0.488 &   0.822 &   0.394 \\
		& FLOT \cite{flot}                  &   0.056 &   0.755 &   0.908 &   0.242 \\
		& SCTN (ours)                              &\bf0.037 &\bf0.873 &\bf0.959 &\bf0.179 \\
		\bottomrule
	\end{tabular}
	\caption{{Comparison with the state-of-the-art} on FlyingThings3D and KITTI. Best results in bold.
		Our proposed model SCTN reaches highest performances in all metrics.}
	\label{tab:comparison}
\end{table*}

\begin{figure*}[t]
	\centering
	\begin{subfigure}{0.236\textwidth}		
		\includegraphics[angle=-0,bb=0 0 2960 2234, width=1\textwidth]{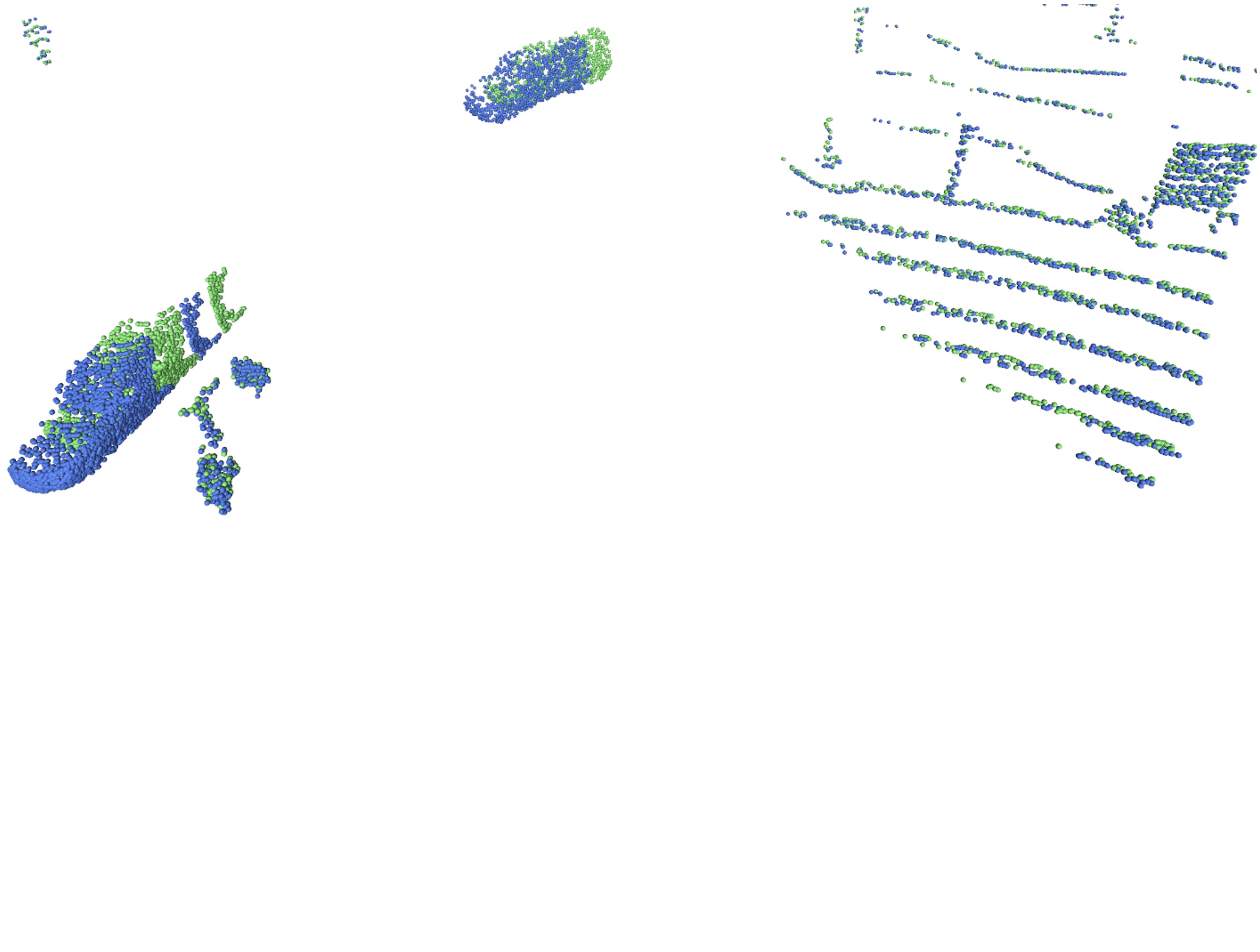}
		\caption{Input point clouds}
	\end{subfigure}
	\begin{subfigure}{0.236\textwidth}		
		\includegraphics[angle=-0,bb=0 0   2960 2234  ,width=1\textwidth]{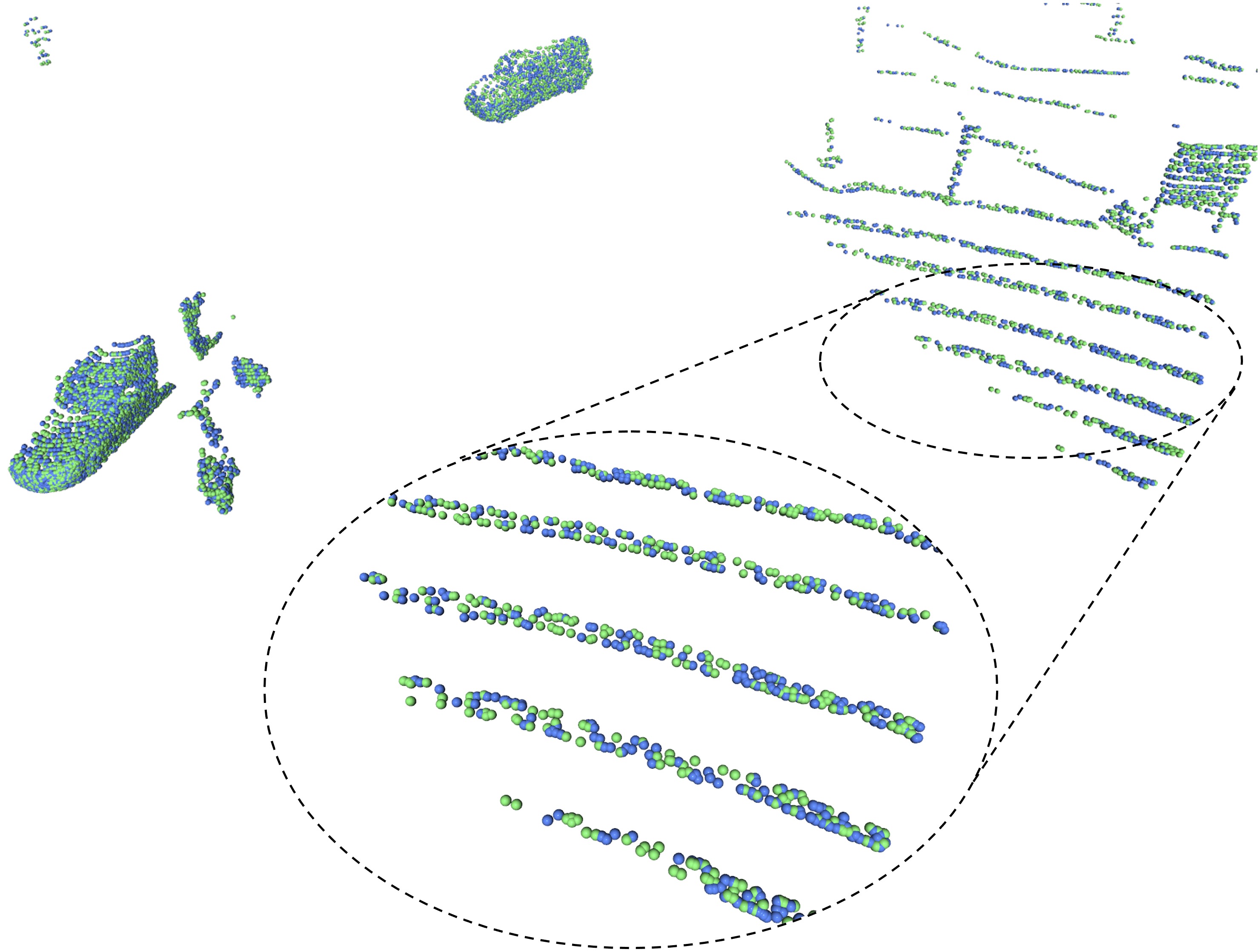}
		\caption{Ground-truth flows}
	\end{subfigure}
	\begin{subfigure}{0.236\textwidth}	
		\includegraphics[angle=-0,bb=0 0    2960 2234 ,width=1\textwidth]{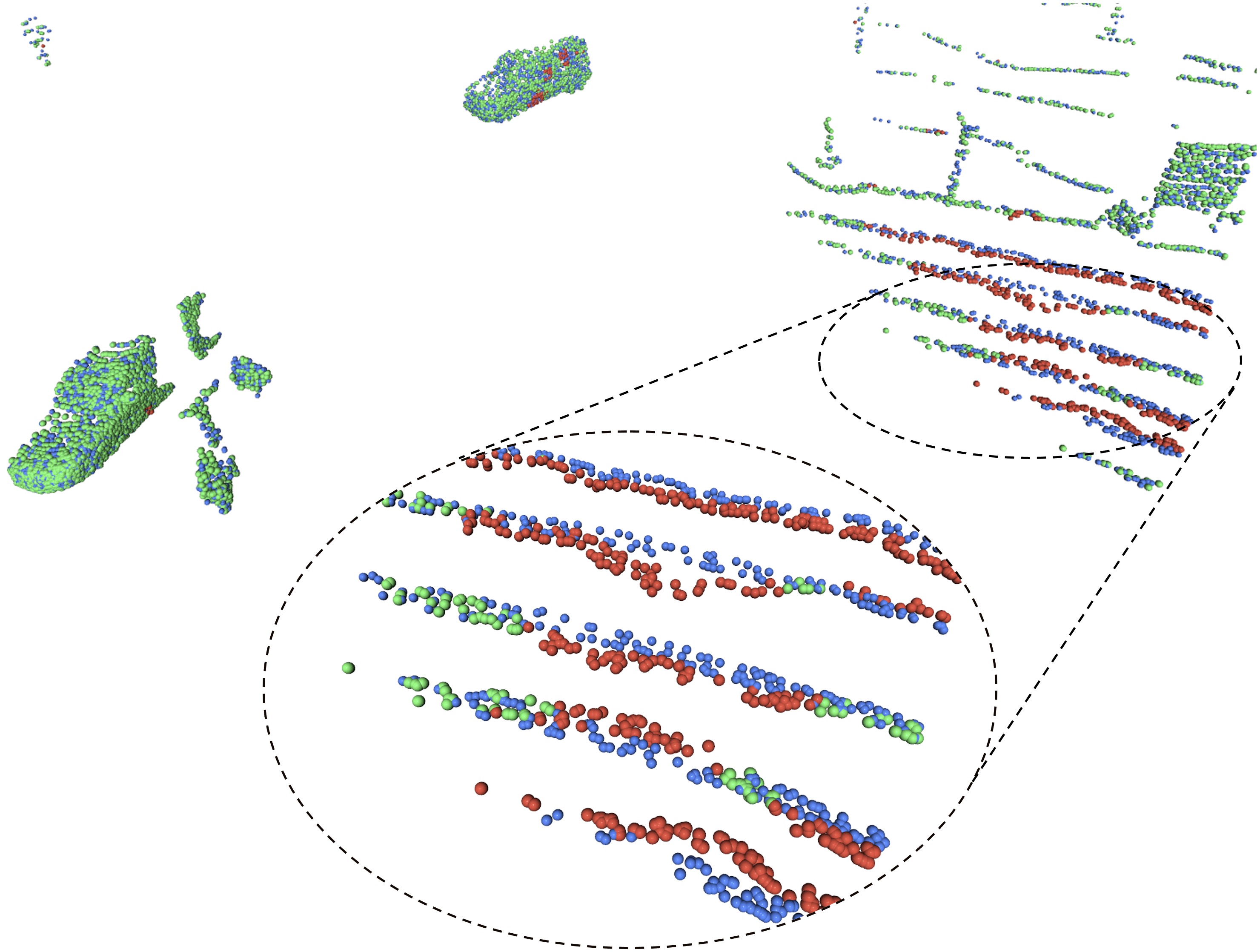}
		\caption{FLOT}
	\end{subfigure}
	\begin{subfigure}{0.236\textwidth}	
		\includegraphics[angle=-0,bb=0 0  2960 2234 ,width=1\textwidth]{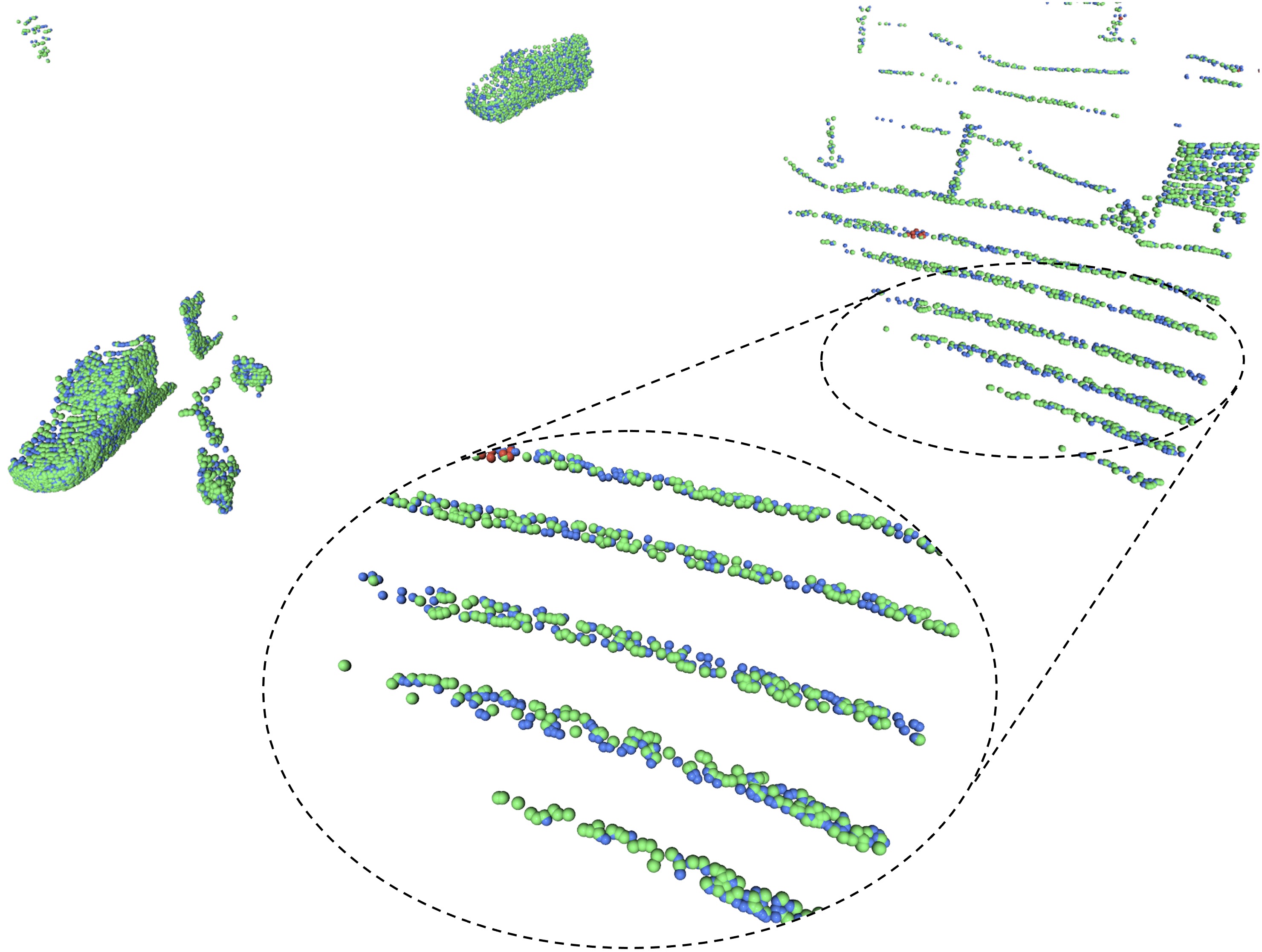}
		\caption{Ours}
	\end{subfigure}
	\caption{Qualitative comparison results.  Green points 	indicate the first point cloud in (a), and 
		blue points indicate the second point cloud in (a)(b)(c)(d). In (b)(c)(d),  green points are the ones in the first point cloud warped by correctly predicted flows, while  red points are the ones warped by incorrect flows (the first point cloud + incorrect scene flow whose EPE3D $>$0.1m).}
	\label{fig:Quali_2}
\end{figure*}

	A naive implementation of the FSC loss would inevitably lead to degenerate cases. In particular, the FSC loss is a product of two objectives: 
	(i) a similarity $s(\mathbf{F}_i,\mathbf{F}_j)$ between the features $\mathbf{F}_i$ and $\mathbf{F}_j$ and 
	(ii) a difference  $\| \mathbf{u}_i-\mathbf{u}_j\|_1$ between their flows.
	The scope of this loss is to train the flows to be similar if they have similar features. However,  to minimize the FSC loss, the model would make the features $\mathbf{F}_j$ and $\mathbf{F}_i$ be orthogonal (\ie $\mathbf{F}_j\cdot\mathbf{F}_i=0$), such that $s(\mathbf{F}_j,\mathbf{F}_i)=0$ (\ie $E^c=0$). Obviously, it is against our aim.
	
	To circumvent this limitation, we propose a stop-gradient for the FSC loss, taking inspiration form recent advances in self-supervised learning~\cite{siasim2020exploring}.
	As illustrated in \Figure{consistencyLoss}, our architecture stops the propagation of the gradient in the branch extracting the feature similarity. By such architecture, our FSC loss  avoids optimizing the features, while optimizing solely the flows similarities $\| \mathbf{u}_j-\mathbf{u}_i\|_1$ for neighboring points with similar features.

	\section{Experiments}
	
	\textbf{Dataset.} We conduct our experiments on two datasets that are widely used to evaluate scene flow. 
	\textit{FlyingThings3D}~\cite{flythings3d} is a large-scale synthetic stereo video datasets, where synthetic objects are selected from ShapeNet \cite{shapenet} and randomly assigned various motions. We generate 3D point clouds and ground truth scene flows with their associated camera parameters and disparities.
	Following the same preprocessing as in \cite{flot,HPLFlowNet,pointpwc}, we randomly sample 8192 points
	and remove points with camera depth greater than 35 m. 
	We use the same $19640/3824$ pairs of point cloud (training/testing) used in the related works~\cite{flot,HPLFlowNet,pointpwc}.
	\textit{KITTI Scene Flow}~\cite{Menze2018JPRS,minkowskiSPC} is a real-world Lidar scan dataset for scene flow estimation from the KITTI autonomous navigation suite. 
	Following the preprocessing of \cite{HPLFlowNet}, we leverage 142 point cloud pairs of 8192 points for testing.
	For a fair comparison, we also remove ground points by discarding points whose height is lower than $-1.4$m, following the setting of existing methods \cite{flot,pointpwc,HPLFlowNet}.

	\textbf{Evaluation Metrics.} To evaluate the performance of our approach, we adopt the standard evaluation metrics used in the related methods \cite{flot,pointpwc}, described as follows:
	The \textit{EPE3D (m)} (3D end-point-error) is calculated by computing the average $\mathit{L}_2$ distance between the predicted and GT scene flow, in meters. This is our main metric.
	The \textit{Acc3DS} is a strict version of the accuracy which estimated as the ratio of points whose EPE3D $<$ 0.05 m or relative error $<$5\%.
	The \textit{Acc3DR} is a relaxed accuracy which is calculated as the ratio of points whose EPE3D $<$0.10m or relative error $<$10\%.
	The \textit{Outliers} is the ratio of points whose EPE3D $>$0.30m or relative error $>$10\%.

	\textbf{Implementation Details.}
	We implement our method in PyTorch \cite{paszke2019pytorch}. We train our method on FlyingThing3D then evaluate on FlyingThing3D and KITTI. 
	We minimize a cumulative loss $E= E^s+\lambda E^c$ with $\lambda=0.30$ a weight that scale the losses. We use the Adam optimizer \cite{kingma2014adam} with an initial learning rate of $10^{-3}$, which is dropped to $10^{-4}$ after the $50^{th}$ epoch. 
	First, we train for $40$ epochs only using  the supervised loss. Then we continue the training for $20$ epochs with both the supervision loss and the FSC loss, for a total on $60$ epochs.  
	We use a voxel size of resolution $0.07$m.

	\textbf{Runtime.} We evaluate the running time of our method. \Table{runtime} reports the evaluated time compared with recent state-of-the-art methods. FLOT \cite{flot} is the most related work to our method, since we both adopt point-wise correlations to generate predicted flows. Our method consumes lower running time than FLOT, although  the transformer module is equipped.

	\subsection{Quantitative Evaluation}
	We compare our approach with recent deep-learning-based methods including FlowNet3D \cite{flownet3d}, HPLFlowNet \cite{HPLFlowNet}, PointPWC \cite{pointpwc} and FLOT \cite{flot}. These methods are state-of-the-art in scene flow estimation from point cloud data and do not leverge any additional labels such as ground-truth ego motions or instance segmentation.

	\textbf{Results on FlyingThings3D.}
	We train and evaluate our model on the FlyThings3D datasets. As shown in \Table{comparison}, our method outperforms all methods in every metrics by a significant margin. 
	It is worth noting that our method obtains an EPE3D metric below $4$cm, with a relative improvement of $26.9\%$ and $35.5\%$ over the most recent methods \FLOT and \PointPWC, respectively.
	The performance shows that our method is effective in predicting flows with high accuracy. 
	
	\textbf{Results on KITTI without Fine-tune.}
	Following the common practice~\cite{flot,pointpwc}, we train our model on FlyingThings3D and directly test the trained model on KITTI Scene Flow dataset, without any fine-tuning, to evaluate the generalization capability of our method. 
	We report in \Table{comparison} the highest accuracy of scene flow estimation on KITTI Scene Flow dataset for our SCTN method.
	Again, we reduce the EPE3D metric below $4$cm, with a 33.9\% relative improvement over \FLOT.
	In the Acc3DS metrics, our method outperforms both \FLOT and \PointPWC by
	$13.5\%$ and $16.6\%$ respectively. 
	This results highlight the capability of our method to generalize well on real-world point cloud  dataset.

\begin{table}[t]
	\centering
	\small
	\begin{tabular}{p{0.6cm} p{0.6 cm}<{\centering}  p{1.3 cm}<{\centering}  p{1.9cm}<{\centering} p{1.8cm}<{\centering} }
		\toprule
		VIFE& PTFE &FSC loss  & EPE3D(m) $\downarrow$ & Acc3DS $\uparrow$    \\

		\midrule
		\ding{51}&           &          &   0.045 &   0.835   \\
		\ding{51}&           &\ding{51} &   0.042 &   0.853    \\
		\ding{51}& \ding{51} &          &   0.040 &   0.863   \\  	
		\ding{51}& \ding{51} &\ding{51} &   \bf0.037 &\bf0.873  \\
		\bottomrule
	\end{tabular}
	\caption{{Ablation for SCTN.} We further analyse the performances of our three components: VIFE, PTFE and FSC loss on KITTI. We highlight in bold the best performances.}
	\label{tab:ablation}
\end{table}
	
	\subsection{Qualitative Evaluation}
	To qualitatively evaluate the quality of our scene flow predictions, we visualize the predicted scene flow and ground-truth one in Figure \ref{fig:Quali_2}.
	Since \FLOT is  most related to our method, we compare the qualitative performances of our SCTN with \FLOT.
	
	Points in a local region from the same object usually have similar ground-truth flows. 
	Yet, FLOT introduces prediction errors in local regions, highlighting the inconsistency in the scene flow predictions.
	For example, FLOT inaccurately predicts scene flow for  some regions in the background, even though those points have similar flows, as shown in Figure \ref{fig:Quali_2}. 
	In contrast, our method is more consistent in the prediction for points in the same object, achieving better performance, \eg for the background objects with complex structure.

	\begin{table}[t]
		\centering
		\small
		\begin{tabular}{c c}
			\toprule
			Method & Runtime (ms)\\
			\midrule
			FLOT \cite{flot} & 389.3 \\
			SCTN (ours) & 242.7\\
			\bottomrule	
		\end{tabular}
		\caption{{Running time comparisons.} The runtime of FLOT and our  SCTN are evaluated on a single GTX2080Ti GPU. We used the official implementation of FLOT.}
		\label{tab:runtime}
	\end{table}
	
	\subsection{Ablation Study}
	To study the roles of the proposed VIFE, PTFE and FSC loss, we ablate each proposed component of our model and evaluate their performance on KITTI datasets. 
	For all the experiments, we follow the same training procedure than in the main results.
	\Table{ablation} reports the evaluation results.
	
	\textbf{VIFE module.} \Table{ablation} shows that our approach with the sole VIEF convolution module already outperforms the state-of-the-art methods listed in \Table{comparison}. Different from existing methods directly applying convolution on original point clouds, our VIFE extracts feature from voxelized point cloud, which reduces the non-uniform density of point cloud, while ensuring that points in a local region have consistent features, to some extent. 
	The results show that such  features are favorable for scene flow estimation.

	\textbf{PTFE module.} Compared with only using VIFE module, adding PTFE improves metrics on KITTI as reported in the third row in \Table{ablation}. 
	For example, EPE3D is improved by 11.1\%, compared with only using the VIEF module.  
	Our PIFE module  explicitly learns point relations, which provides rich contextual information and helps to match corresponding points even for objects with complex structures.

	\textbf{FSC loss.} \Table{ablation} shows that adding the FSC loss helps to achieve better scene flow estimation on KITTI.  Our FSC loss improves the generalization capability of our method.

	\section{Conclusion}
	We present a Sparse Convolution-Transformer Network (SCTN) for scene flow estimation.  Our SCTN leverages the VIFE module to transfer irregular point cloud into locally smooth
	flow features for estimating spatially consistent motions in local regions. Our PTFE module learns rich contextual information via explicitly modeling point relations, which is
	helpful for matching corresponding points and benefits scene
	flow estimation.  A novel FSC loss is also proposed for training SCTN, improving the generalization ability of our method.
	Our approach achieves state-of-the-art performances on FlyingThings3D and KITTI  datasets.

	\section{Acknowledgments}
	This work was supported by the King Abdullah University of Science and Technology (KAUST) Office of Sponsored Research through the Visual Computing Center (VCC) funding. 
	We thank Hani Itani for his constructive suggestions and help.
	
	{\small
	\bibliographystyle{ieee_fullname}
	\bibliography{flow1}
}
	
	\clearpage

	\section{Appendix}

\input{sections/7_appendix}

\end{document}

%% file: sections/7_appendix.tex
\begin{figure*}[t]
	\centering
	\subcaptionbox{Input point clouds}{
		\begin{minipage}[t]{0.23\textwidth}	
			\includegraphics[angle=-11,bb=0 0  711 581 ,width=1\textwidth]{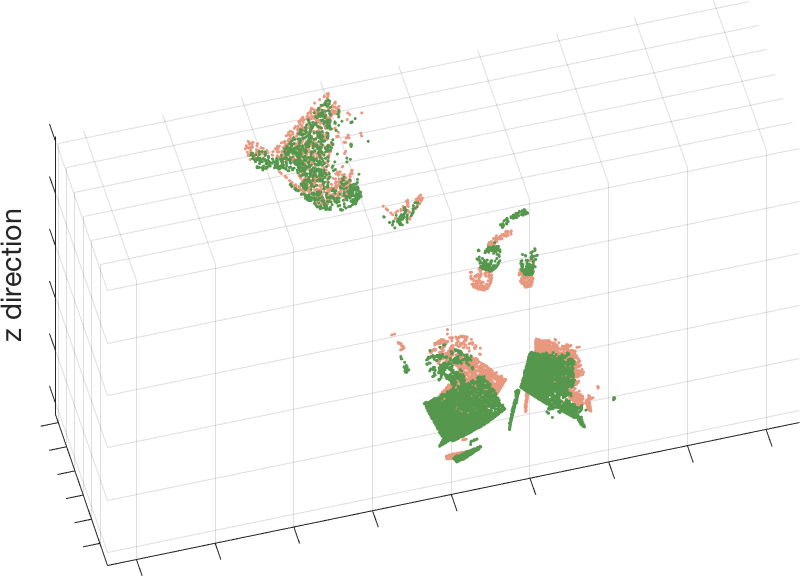}
			\includegraphics[angle=-11,bb=0 0  711 581 ,width=1\textwidth]{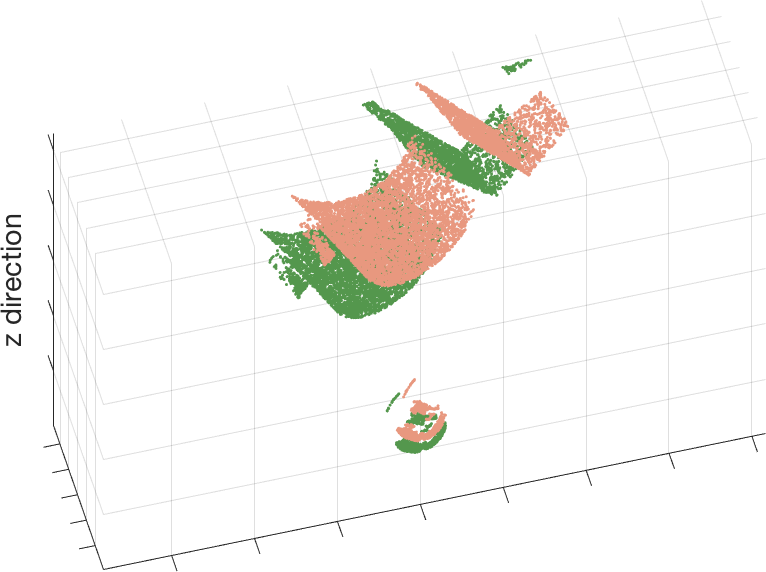}
			\includegraphics[angle=-11,bb=0 0  711 581 ,width=1\textwidth]{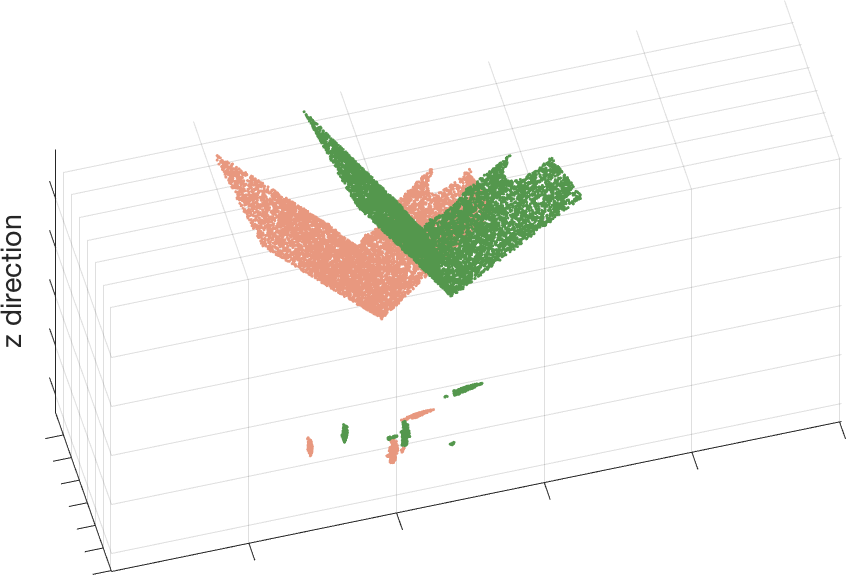}
		\end{minipage}
	}
	\subcaptionbox{Ground-truth flows}{
		\begin{minipage}[t]{0.23\textwidth}	
			\includegraphics[angle=-11,bb=0 0  711 581 ,width=1\textwidth]{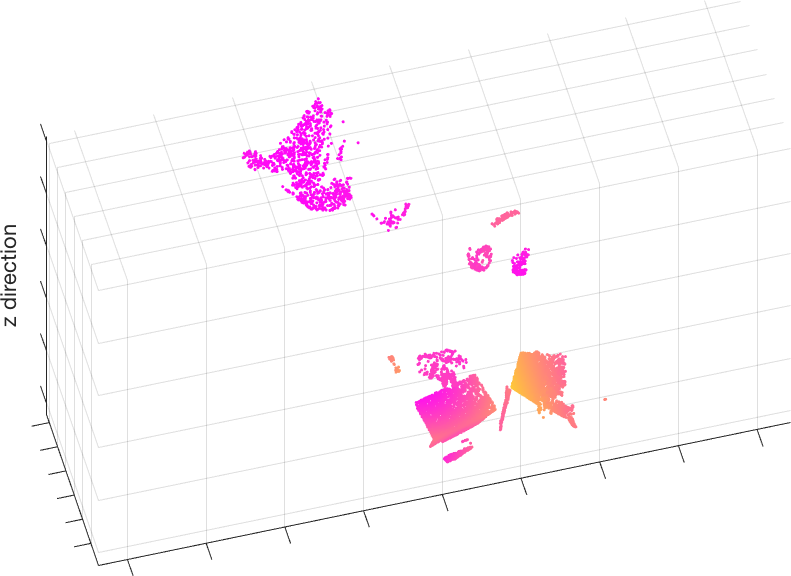}
			\includegraphics[angle=-11,bb=0 0  711 581 ,width=1\textwidth]{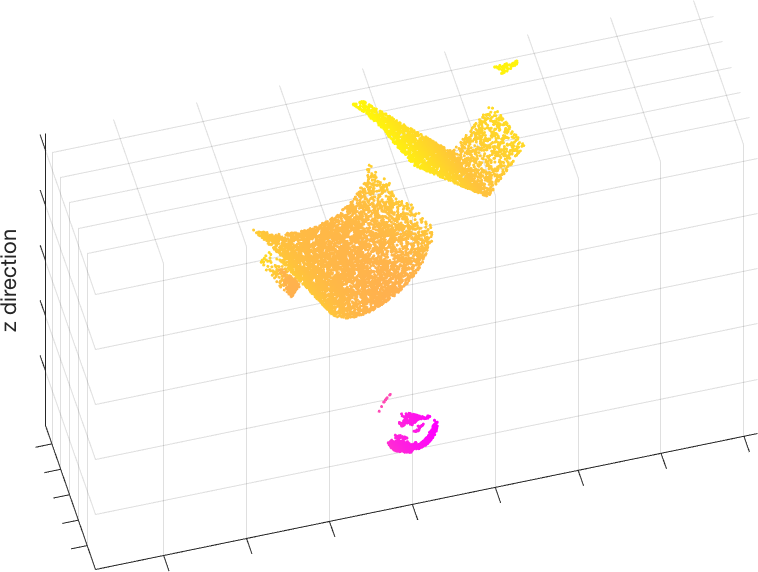}
			\includegraphics[angle=-11,bb=0 0  711 581 ,width=1\textwidth]{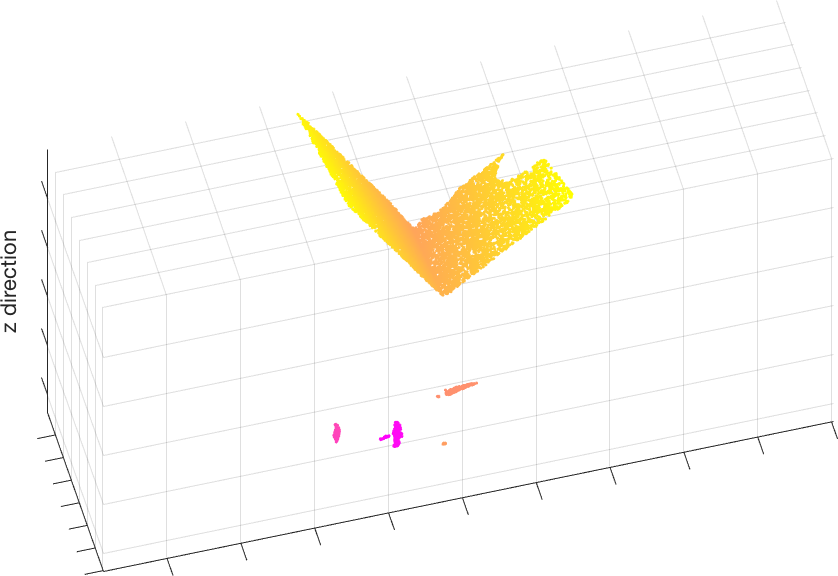}
		\end{minipage}
	}
	\subcaptionbox{FLOT}{
		\begin{minipage}[t]{0.23\textwidth}	
			\includegraphics[angle=-11,bb=0 0  711 581 ,width=1\textwidth]{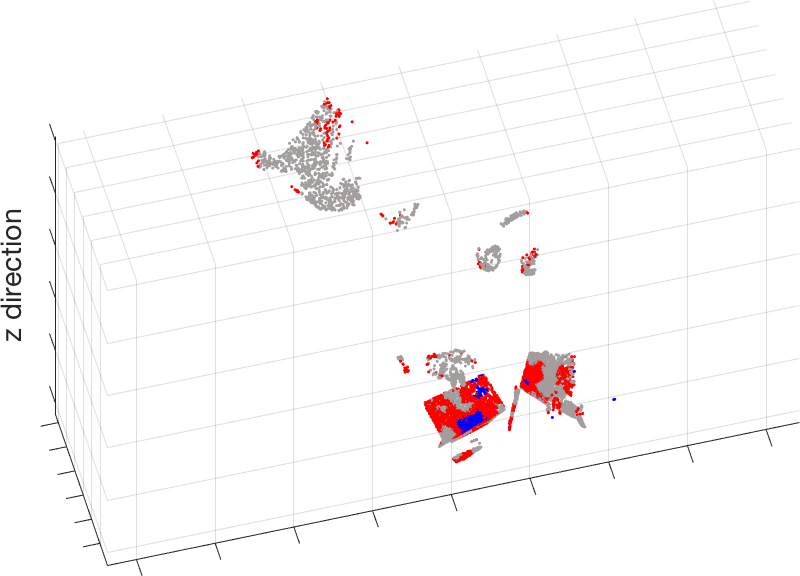}
			\includegraphics[angle=-11,bb=0 0  711 581 ,width=1\textwidth]{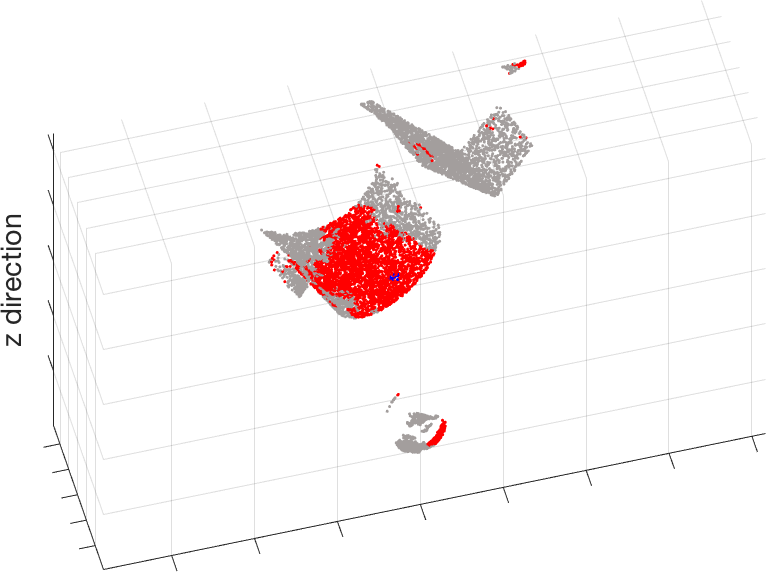}
			\includegraphics[angle=-11,bb=0 0  711 581 ,width=1\textwidth]{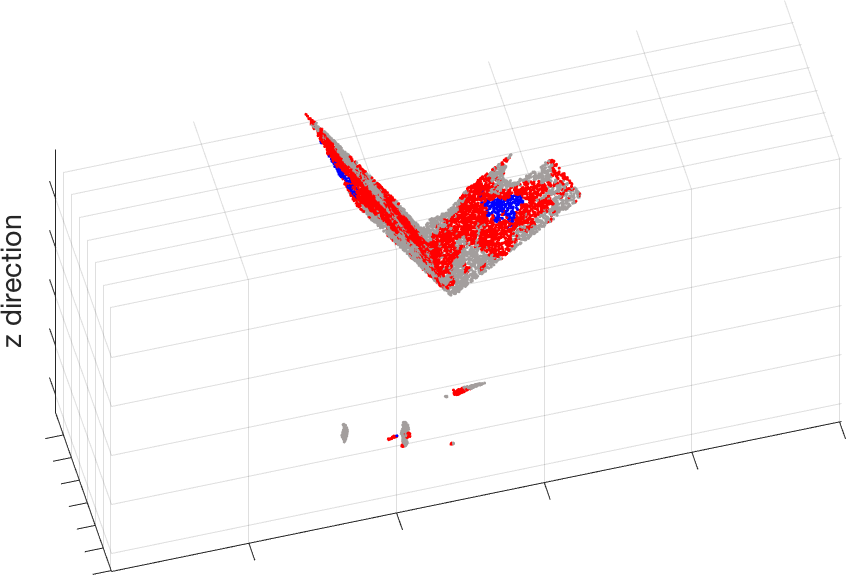}
		\end{minipage}
	}
	\subcaptionbox{Ours}{
		\begin{minipage}[t]{0.23\textwidth}	
			\includegraphics[angle=-11,bb=0 0  711 581 ,width=1\textwidth]{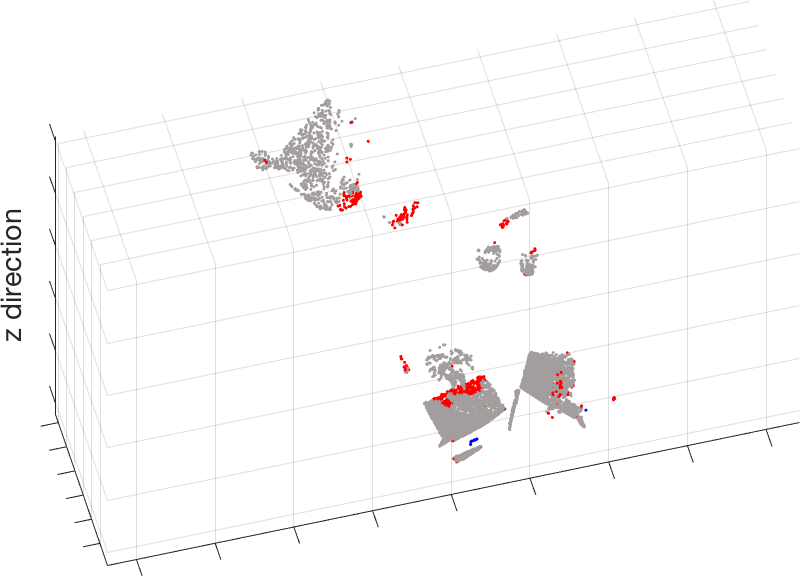}
			\includegraphics[angle=-11,bb=0 0  711 581 ,width=1\textwidth]{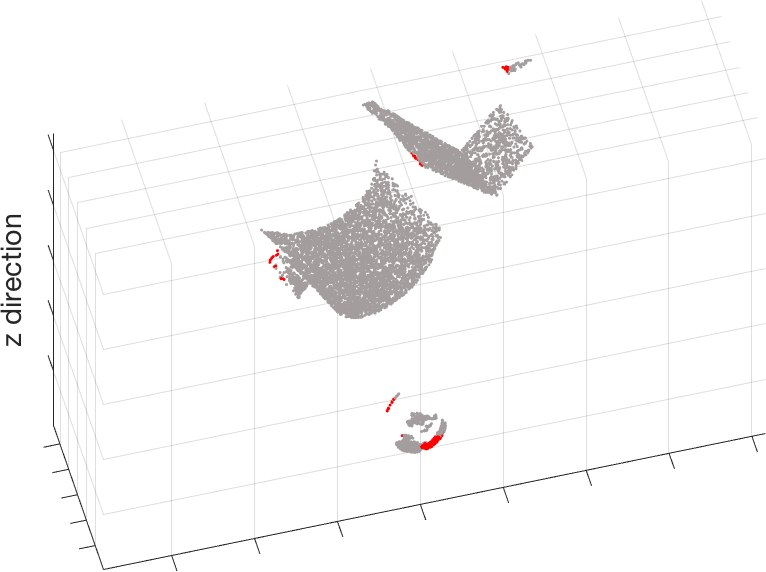}
			\includegraphics[angle=-11,bb=0 0  711 581 ,width=1\textwidth]{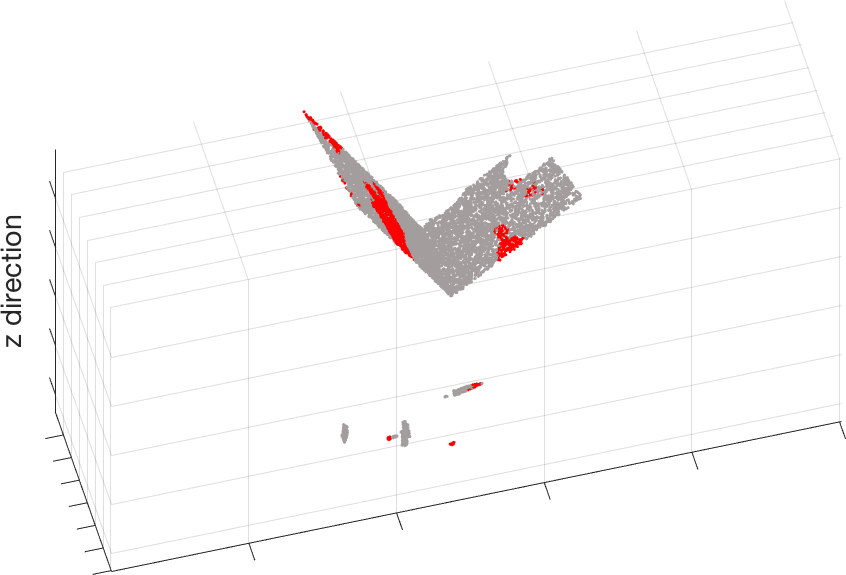}
		\end{minipage}
	}
	\caption{\textbf{Qualitative comparisons between FLOT \cite{flot} and our method on Flythings3D  dataset}. Orange and green indicates the first and second point clouds in (a).  The similar  color indicates the point cloud has the similar flows in  (b). Gray, {\color{red}red} and {\color{blue}blue} color indicate small, medium and large errors in (c)(d).}
	\label{fig:Quali_our_fly46}
	\vspace*{-14pt}
\end{figure*}

\begin{figure*}[t]
	\centering
	\subcaptionbox{Input point clouds}{
		\begin{minipage}[t]{0.23\textwidth}	
			\includegraphics[angle=-11,bb=0 0  711 581 ,width=1\textwidth]{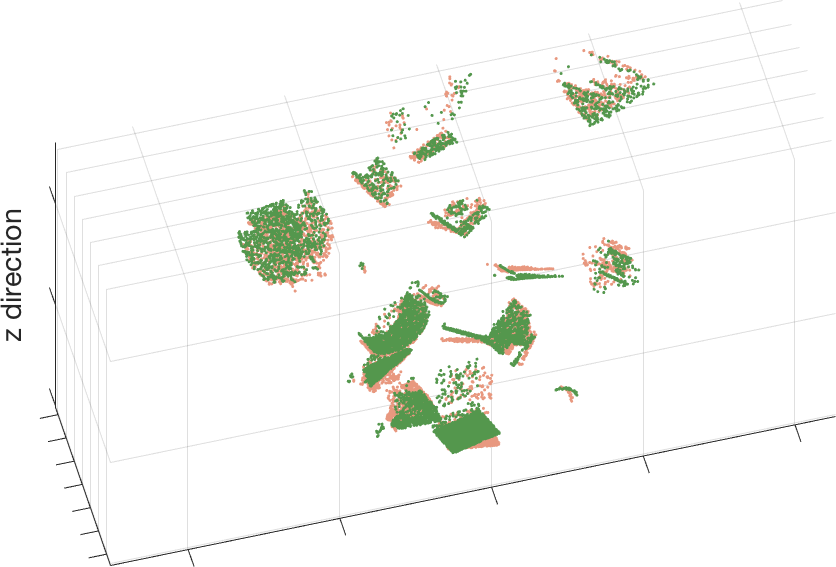}
			\includegraphics[angle=-11,bb=0 0  711 581 ,width=1\textwidth]{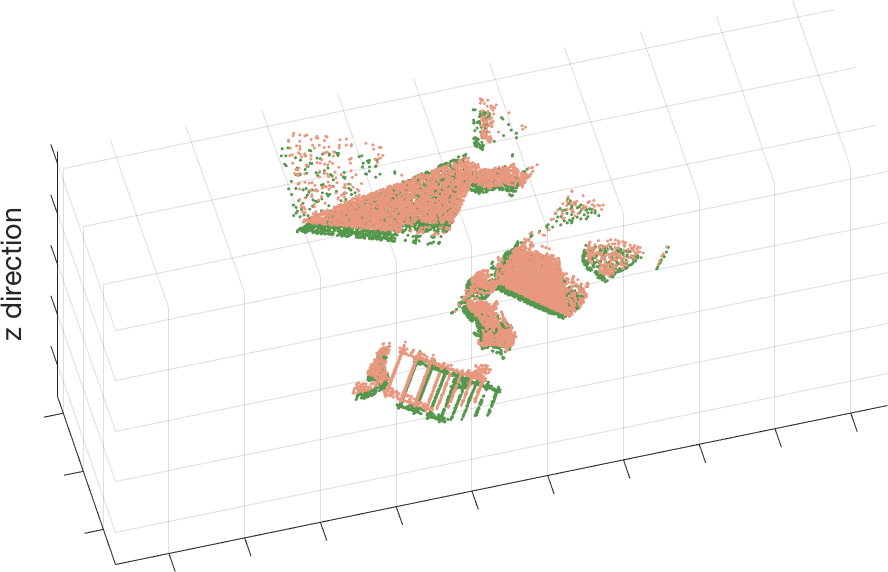}
			\includegraphics[angle=-11,bb=0 0  711 581 ,width=1\textwidth]{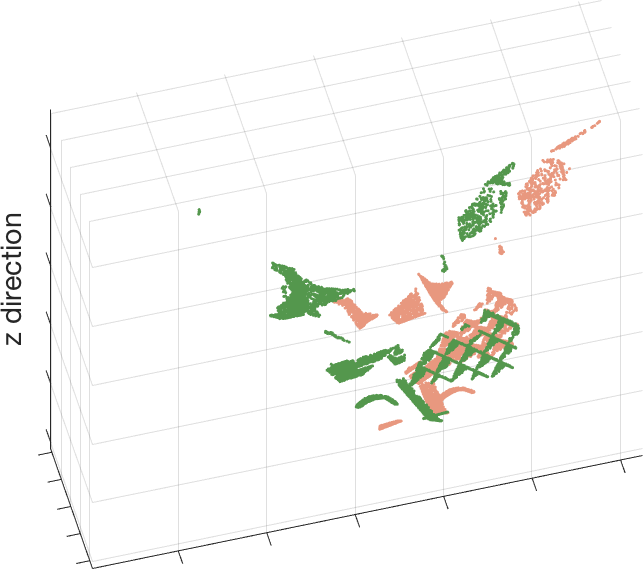}
		\end{minipage}
	}
	\subcaptionbox{Ground-truth flows}{
		\begin{minipage}[t]{0.23\textwidth}	
			\includegraphics[angle=-11,bb=0 0  711 581 ,width=1\textwidth]{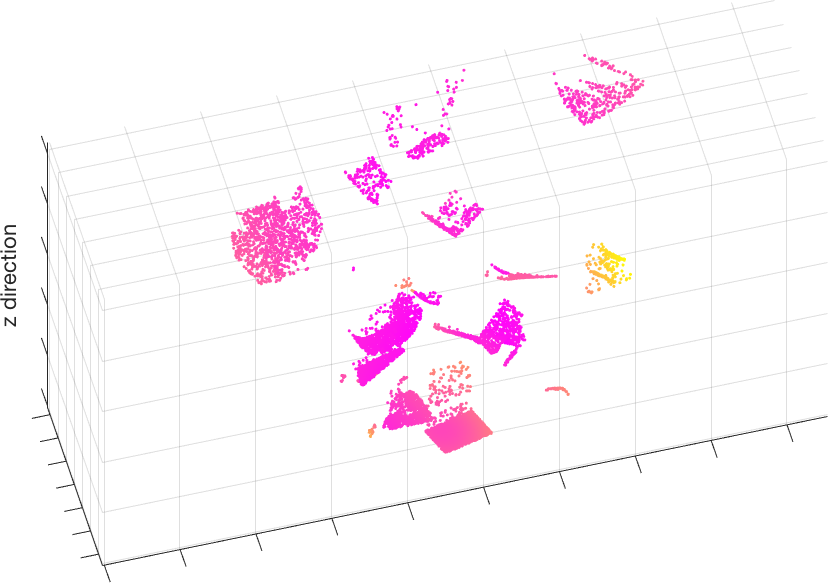}
			\includegraphics[angle=-11,bb=0 0  711 581 ,width=1\textwidth]{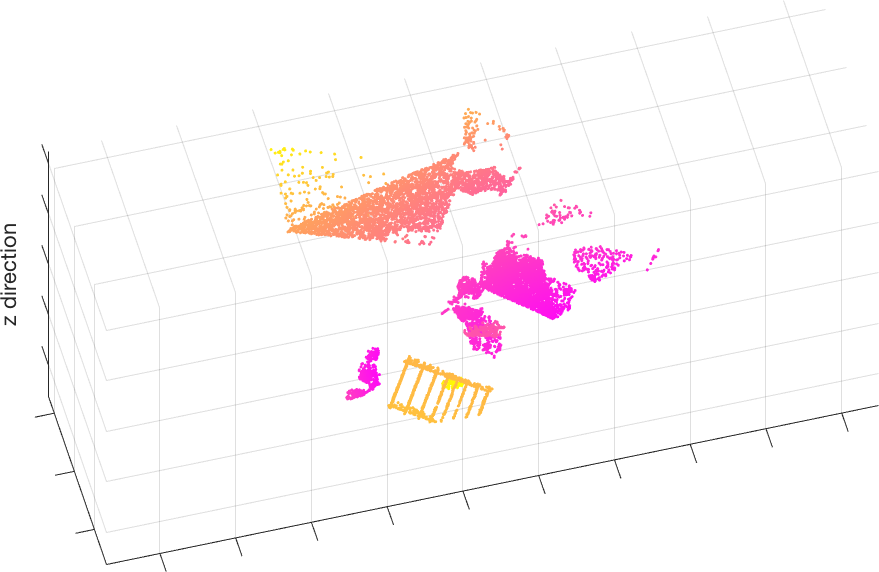}
			\includegraphics[angle=-11,bb=0 0  711 581 ,width=1\textwidth]{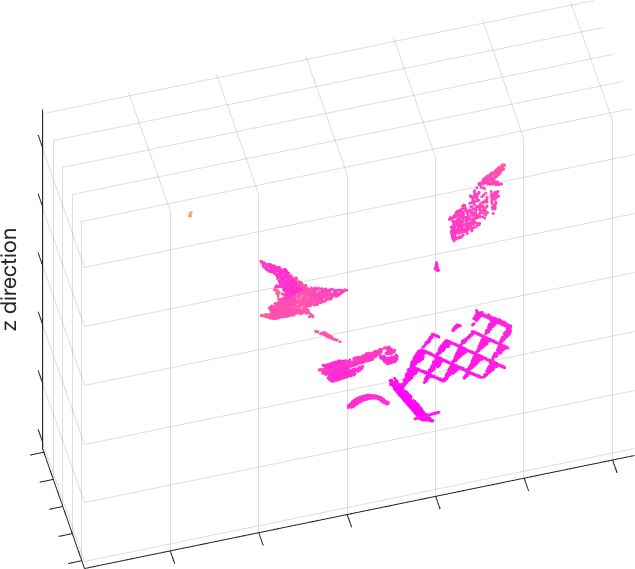}
		\end{minipage}
	}
	\subcaptionbox{FLOT}{
		\begin{minipage}[t]{0.23\textwidth}	
			\includegraphics[angle=-11,bb=0 0  711 581 ,width=1\textwidth]{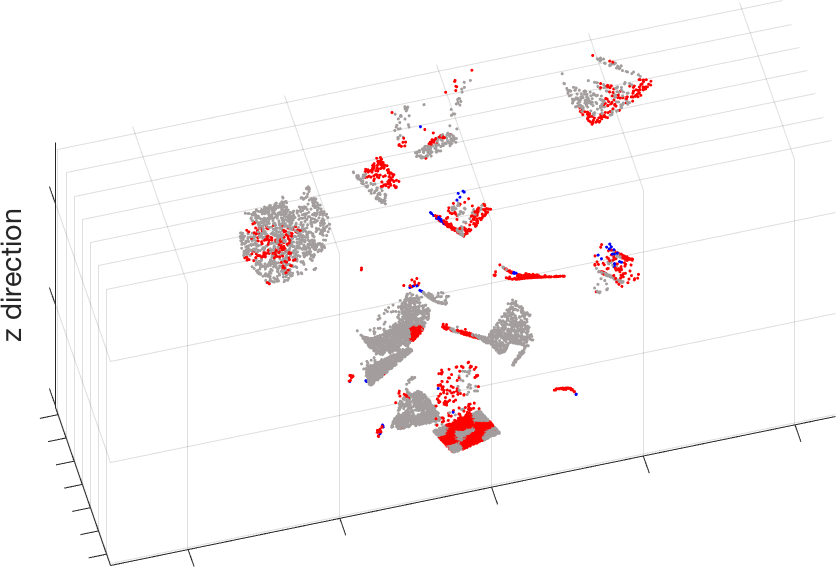}
			\includegraphics[angle=-11,bb=0 0  711 581 ,width=1\textwidth]{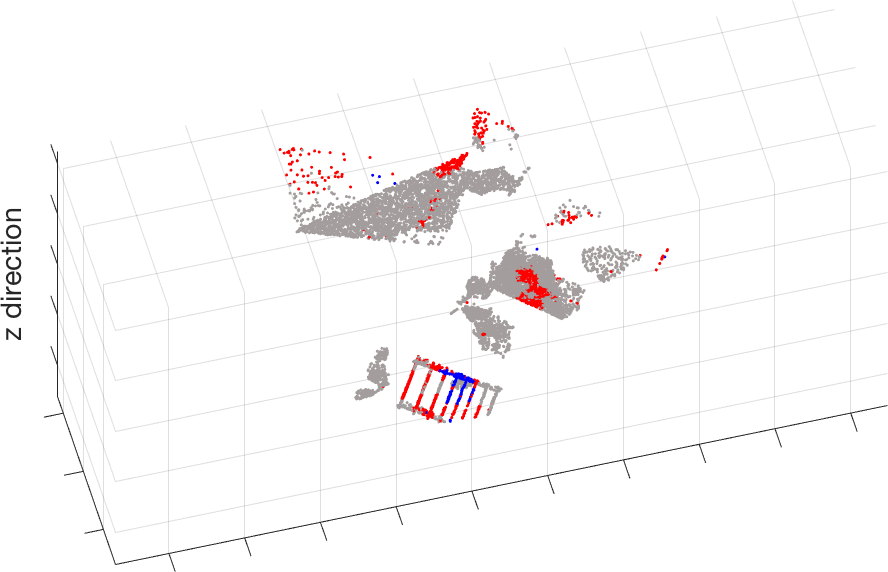}
			\includegraphics[angle=-11,bb=0 0  711 581 ,width=1\textwidth]{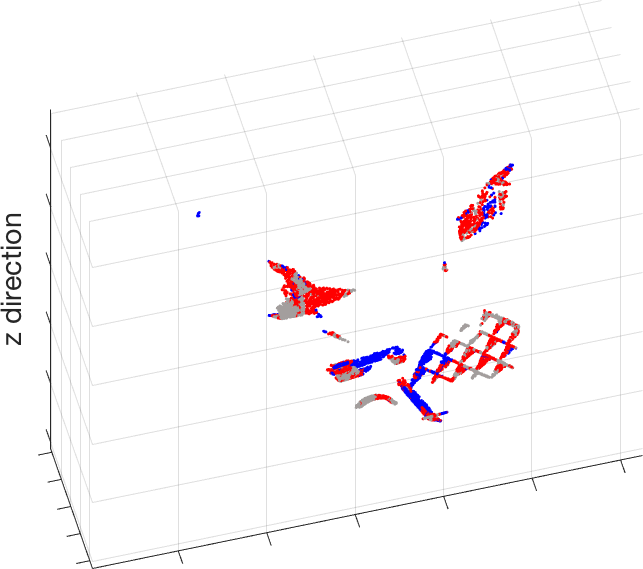}
		\end{minipage}
	}
	\subcaptionbox{Ours}{
		\begin{minipage}[t]{0.23\textwidth}	
			\includegraphics[angle=-11,bb=0 0  711 581 ,width=1\textwidth]{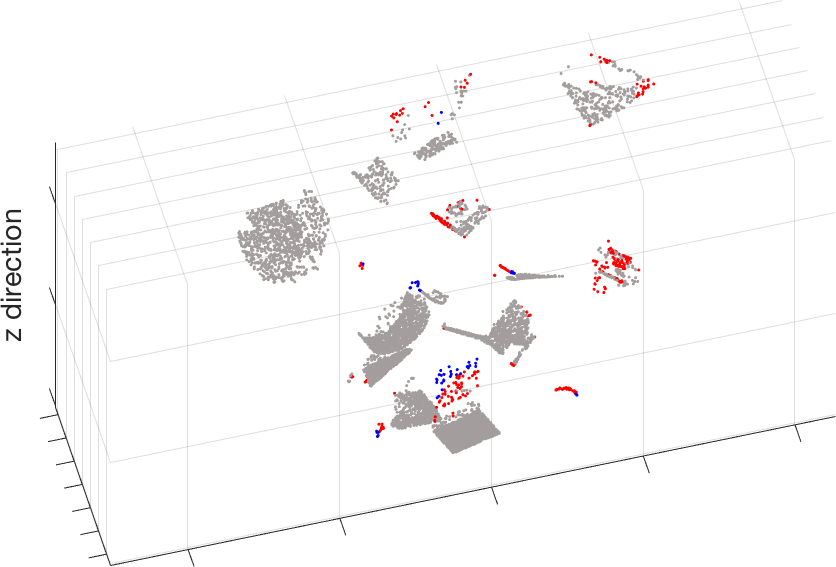}
			\includegraphics[angle=-11,bb=0 0  711 581 ,width=1\textwidth]{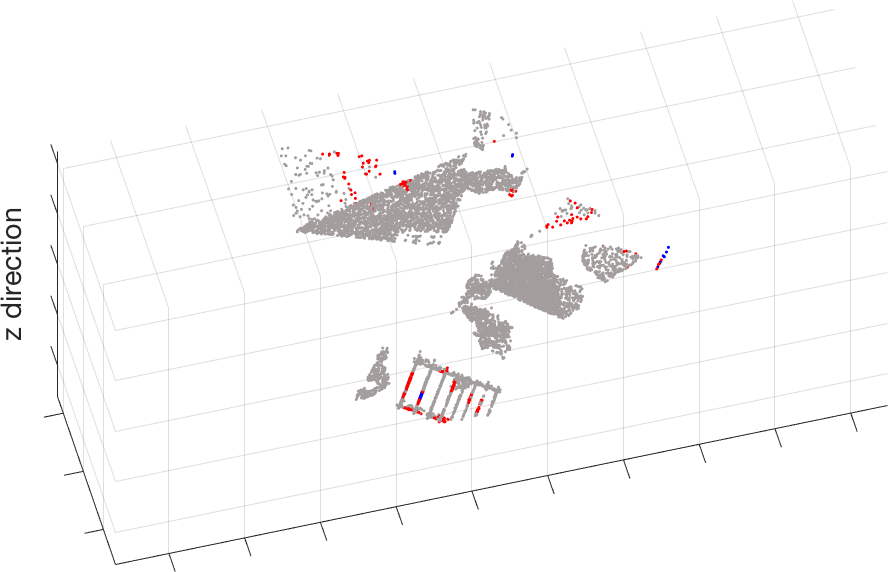}
			\includegraphics[angle=-11,bb=0 0  711 581 ,width=1\textwidth]{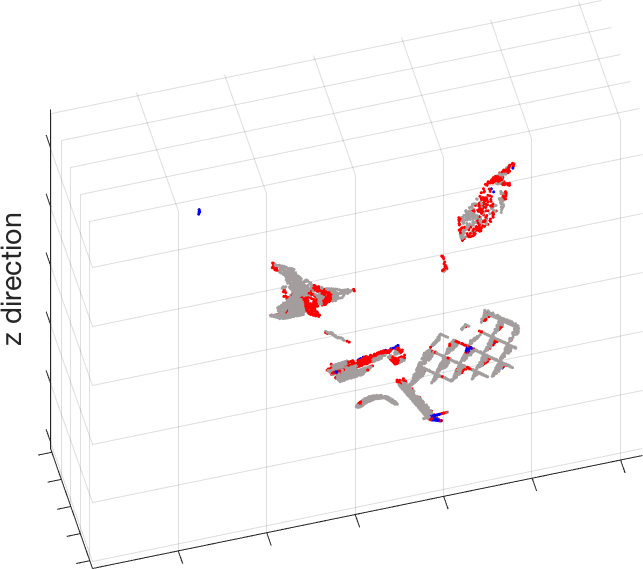}
		\end{minipage}
	}
	\caption{\textbf{Qualitative comparisons between FLOT \cite{flot} and our method on Flythings3D  dataset}. Orange and green indicates the first and second point clouds in (a).  The similar  color indicates the point cloud has the similar flows in  (b). Gray, {\color{red}red} and {\color{blue}blue} color indicate small, medium and large errors in (c)(d).}
	\label{fig:Quali_our_fly230}
	\vspace*{-14pt}
\end{figure*}

\begin{figure*}[t]
	
	\centering
	\subcaptionbox{Input point clouds}{
		\begin{minipage}[t]{0.23\textwidth}	
			\includegraphics[angle=-11,width=1\textwidth]{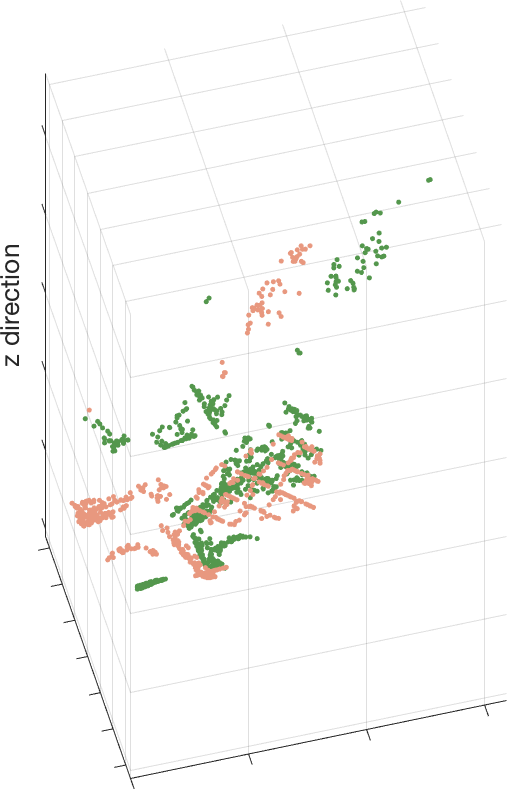}
			\includegraphics[angle=-11,width=1\textwidth]{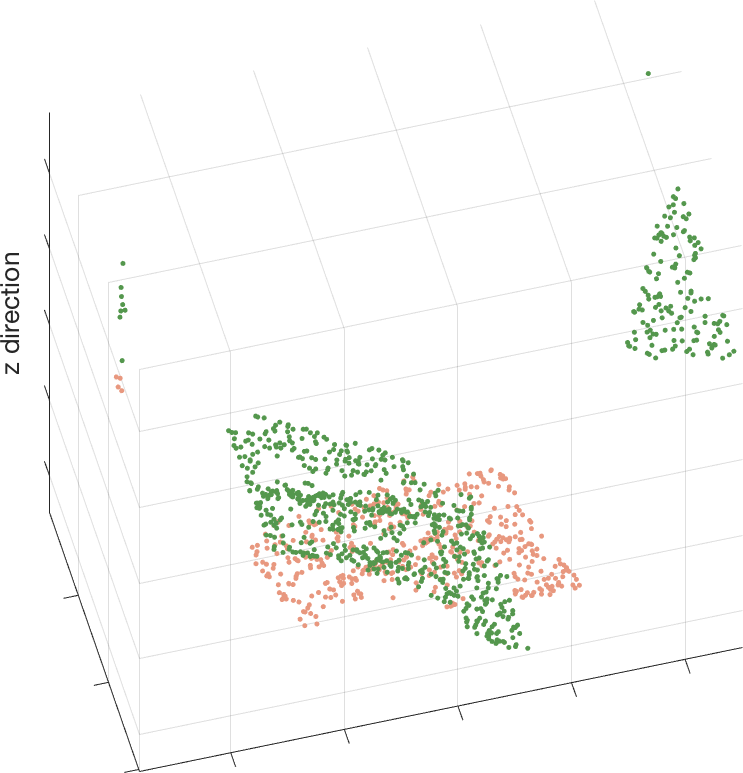}
		\end{minipage}
	}
	\subcaptionbox{warped point clouds   \newline by ground-truth flows}{
		\begin{minipage}[t]{0.23\textwidth}	
			\includegraphics[angle=-11, width=1\textwidth]{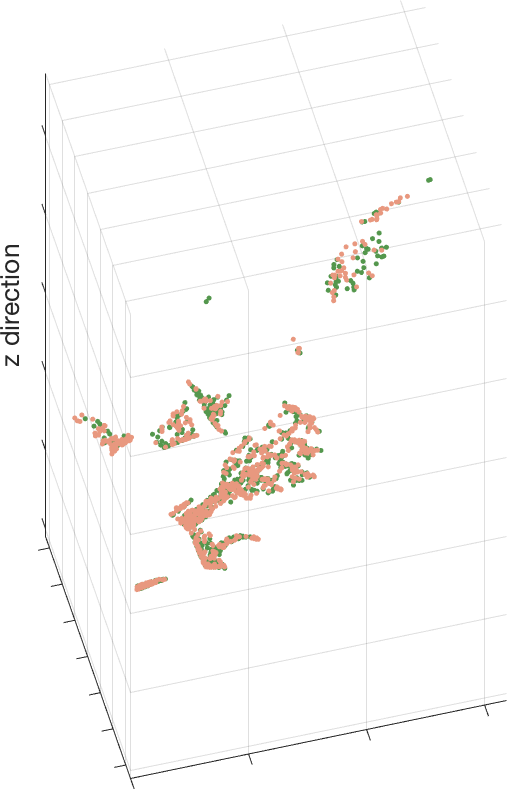}
			\includegraphics[angle=-11, width=1\textwidth]{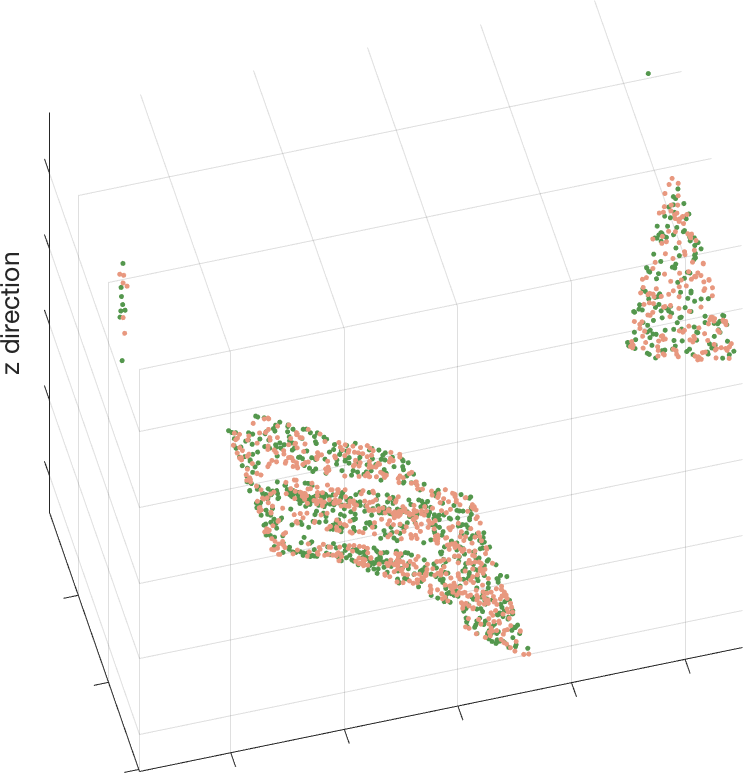}
		\end{minipage}
	}
	\subcaptionbox{warped point clouds  \newline  by predicted flows of  FLOT}{
		\begin{minipage}[t]{0.23\textwidth}	
			\includegraphics[angle=-11,width=1\textwidth]{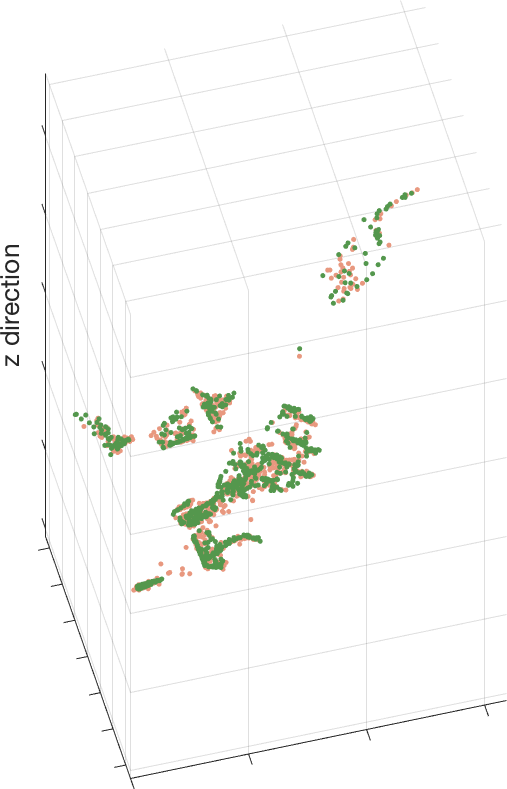}
			\includegraphics[angle=-11,width=1\textwidth]{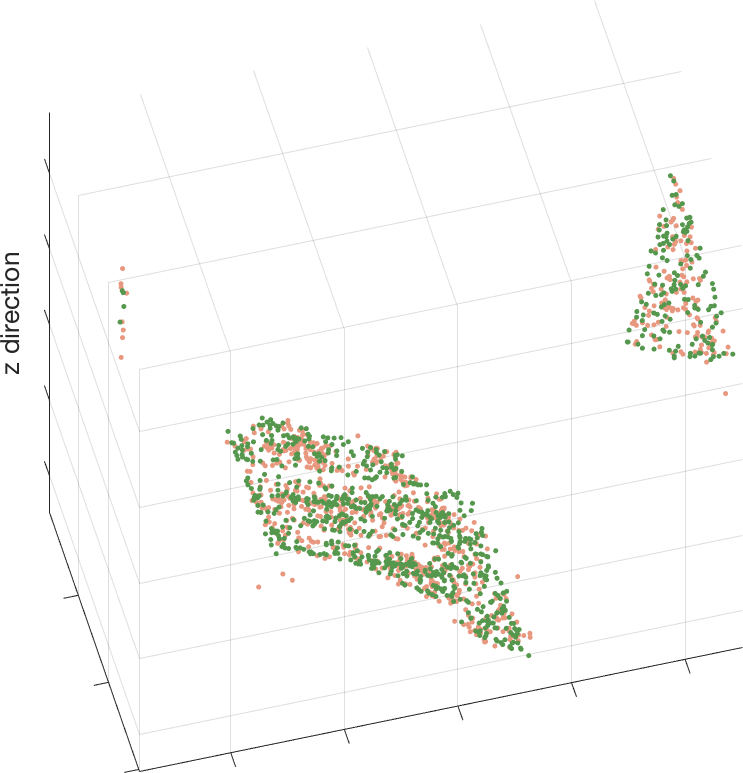}
		\end{minipage}
	}
	\subcaptionbox{warped point clouds  \newline  by our predicted flows}{
		\begin{minipage}[t]{0.23\textwidth}	
			\includegraphics[angle=-11,width=1\textwidth]{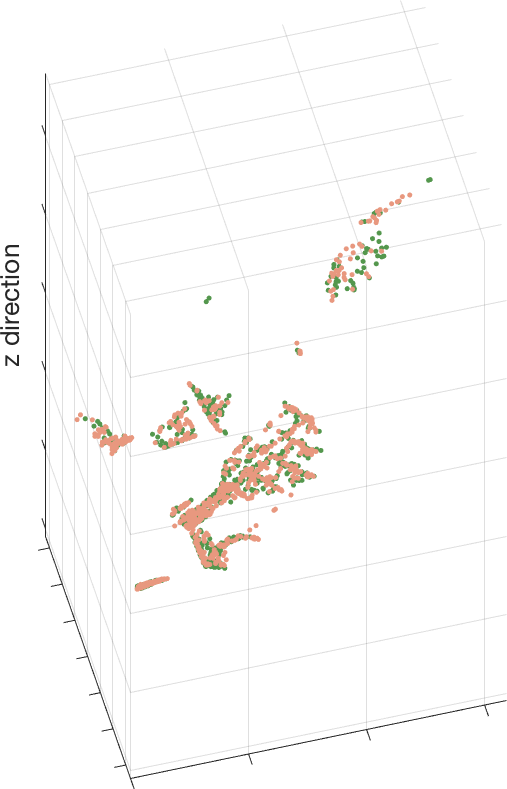}
			\includegraphics[angle=-11,width=1\textwidth]{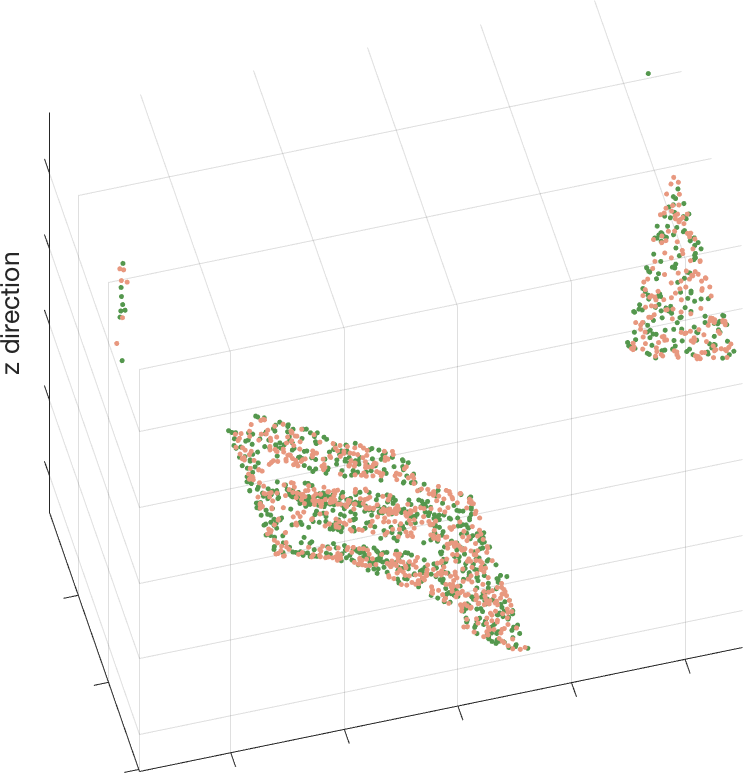}
		\end{minipage}
	}
	\caption{\textbf{Qualitative comparisons between FLOT \cite{flot} and our method on Flythings3D  dataset}. Orange and green indicates the first and second point clouds in (a)(b)(c)(d). The first point cloud is warped by ground-truth flows in (b),  and is  warped by predicted flows of FLOT  in (c) and  our method in (d), respectively.  Compared with FLOT, the warped first point cloud by our predicted flows is  more similar to that by ground-truth flows,  indicating better estimation accuracy of our method.}
	\label{fig:Quali_our_fly206warp}
	\vspace*{-14pt}
\end{figure*}

\subsection{Qualitative Results on FlyingThings3D}
We provide qualitative results of our method on FlyingThings3D dataset.

Figure \ref{fig:Quali_our_fly46}  shows three examples where point clouds are with large motions (see large position differences between orange and greed point cloud in Figure  \ref{fig:Quali_our_fly46}{\color{red}a}). FLOT introduces medium errors in large regions of objects especially in uniform regions. In contrast, our method better predicts flows, since our transformer captures rich context information for finding correspondences and FSC loss adaptively enforces flow consistency.

Figure \ref{fig:Quali_our_fly230}  shows three examples where point clouds containing complex scene or objects with complex structure. Our method achieves better performance than FLOT. Our transformer can learn meaningful structure information via explicitly modeling point relation. 

In addition, we warp the first point cloud $\mathcal{P}$ using the predicted scene flow and visualize the warped point cloud, like FLOT \cite{flot} and PointPWC \cite{pointpwc} did. If the warped point cloud using predicted flows is similar to that using ground-truth flows, the predicted flows are of high accurate.
As shown in Figure \ref{fig:Quali_our_fly206warp}, since our method predicts higher accurate flows than FLOT, the warped point cloud using our predicted flows is more similar to that using ground-truth flows.